\documentclass{article}
\usepackage{spconf,amsmath,epsfig}

\usepackage{subcaption}
\usepackage{amsmath}
\usepackage{amssymb}
\usepackage[capitalize]{cleveref} 
\newcommand{\signedDist}{\widehat{D}_\mathbf{p}}
\newcommand{\radiance}{C_{\mathbf{p},  \mathbf{d} }^{\,i}}
\newcommand{\rayColor}{\widehat{C}_{\mathbf{r}}^{\,i}}
\usepackage{multirow}
\usepackage{xspace}
\usepackage{color}
\usepackage{url}
\usepackage[sectionbib]{bibunits}

\newcommand{\infusionsurf}{InFusionSurf\xspace}
\newcommand{\tsdffusion}{TSDF Fusion\xspace}
\newcommand{\ceil}[1]{\lceil {#1} \rceil}
\newcommand{\perframe}{per-frame intrinsic refinement\xspace}
\newcommand{\Perframe}{Per-frame intrinsic refinement\xspace}
\newcommand{\ipdf}{image-plane deformation field\xspace}
\newcommand{\projctPageLink}{\url{https://rokit-healthcare.github.io/InFusionSurf}}

\let\OLDthebibliography\thebibliography
\renewcommand\thebibliography[1]{
  \OLDthebibliography{#1}
  \setlength{\parskip}{0pt}
  \setlength{\itemsep}{0pt plus 0.3ex}
}

\defaultbibliographystyle{IEEEbib}
\defaultbibliography{main}

\ninept

\begin{document}\sloppy
\bibliographyunit[\title]

\title{InFusionSurf: Refining Neural RGB-D Surface Reconstruction Using Per-Frame Intrinsic Refinement and TSDF Fusion Prior Learning}
%
\name{Seunghwan Lee, Gwanmo Park, Hyewon Son, Jiwon Ryu, Han Joo Chae}
\address{ROKIT Healthcare, Inc.}

\maketitle

\begin{abstract}
    We introduce InFusionSurf, an innovative enhancement for neural radiance field (NeRF) frameworks in 3D surface reconstruction using RGB-D video frames. 
    Building upon previous methods that have employed feature encoding to improve optimization speed, we further improve the reconstruction quality with minimal impact on optimization time by refining depth information. 
    InFusionSurf addresses camera motion-induced blurs in each depth frame through a per-frame intrinsic refinement scheme. It incorporates the truncated signed distance field (TSDF) Fusion, a classical real-time 3D surface reconstruction method, as a pretraining tool for the feature grid, enhancing reconstruction details and training speed. Comparative quantitative and qualitative analyses show that InFusionSurf reconstructs scenes with high accuracy while maintaining optimization efficiency. The effectiveness of our intrinsic refinement and TSDF Fusion-based pretraining is further validated through an ablation study.
\end{abstract}

\section{Introduction}

The integration of a depth-measurement-based implicit surface representation into the volume rendering of neural radiance fields (NeRF)~\cite{nerf} by Neural RGB-D~\cite{NeuralRGBD} has significantly improved the quality of geometry estimation in 3D surface reconstruction. 
However, like many NeRF variants, Neural RGB-D suffers from lengthy optimization times for new scenes, taking 9 to 13 hours depending on the scene size. 
Although recent advancements in explicit representations~\cite{DVGO, InstantNGP, GOSurf} have notably reduced these optimization times, further improvements are necessary for their practical use in real-world applications demanding quick and accurate 3D surface reconstruction.

Additionally, commercial image-capturing devices often introduce distortions like motion blur, defocus blur, and rolling shutter effects in video frames, challenging NeRF methods in producing sharp images from these blurry inputs. 
While some NeRF-integrated deblurring approaches~\cite{dpnerf, BAD_nerf} have been developed for color frames, they are less effective for depth-dependent 3D reconstruction since camera motion blurs on depth frames, differing from those in color frames, have greater impact on the reconstruction results.

In this work, we introduce \emph{\infusionsurf}\footnote{\projctPageLink}, a NeRF-style RGB-D 3D surface reconstruction framework that refines depth information to enhance reconstruction quality with minimal impact on optimization time.
Our \perframe scheme employs explicit parameters to efficiently optimize ray casting directions, addressing camera motion blurs in depth video frames.
Additionally, \infusionsurf leverages the truncated signed distance field (TSDF) Fusion~\cite{SDF}, a classical real-time 3D surface reconstruction method, as a pretraining step, to give the model a head-start. 
We demonstrate \infusionsurf's ability to reconstruct scenes accurately and efficiently compared to prior works, Neural RGB-D~\cite{NeuralRGBD} and GO-Surf~\cite{GOSurf}. We further validate the effectiveness of \tsdffusion prior learning and \perframe techniques through an ablation study.

\begin{figure*}[t!]
    \centering
    \includegraphics[width=0.8\textwidth]{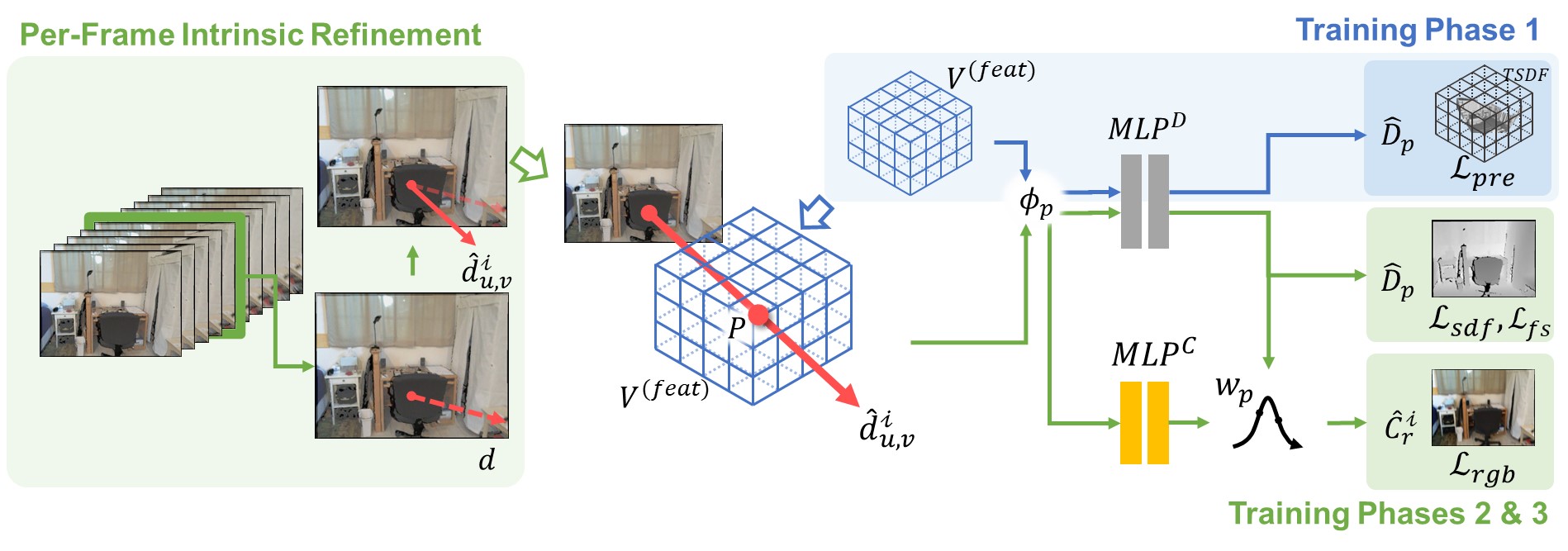}
    \caption{Our method proposes per-frame intrinsic refinement and classical TSDF Fusion prior learning schemes for high-quality 3D surface reconstruction with minimal impact on optimization time. We adopt the Neural RGB-D method, revised with a dense feature grid and shallow MLPs. Our per-frame intrinsic refinement scheme compensates for the frame-specific distortion effects caused by the camera motion. The first phase of the training learns geometric prior using the TSDF Fusion algorithm and the later phases adopt a progressive learning technique.}
    \label{fig:teaser}
\end{figure*}
\section{Related work}

\subsection{Classical 3D reconstruction}
Variants of \tsdffusion method~\cite{SDF} have been commonly used for reconstructing 3D surfaces from depth images~\cite{survey}.
Over the years, numerous improvements have been implemented, ranging from real-time applications~\cite{KinectFusion} to enhanced reconstruction quality~\cite{BundleFusion}.

Despite their suboptimal quality, we found that leveraging the efficient \tsdffusion output as a geometric prior improves the reconstruction quality while accelerating the convergence speed.

\subsection{Neural radiance field and depth}
Various attempts have been made to adapt the neural radiance field representation and volume rendering scheme~\cite{nerf} to depth images.
Some methods incorporated depth priors for better novel view synthesis~\cite{depthNeRF, DONeRF}, while others used implicit neural representations for 3D surface reconstruction~\cite{NeuralRGBD, GOSurf}.

Given the long optimization time of NeRF and its variants, methods for quicker convergence have been proposed~\cite{DVGO, InstantNGP}.
GO-Surf~\cite{GOSurf} combines a multi-resolution feature grid with a hybrid volume rendering scheme akin to Neural RGB-D for faster optimization speed.

Our approach adopts Neural RGB-D~\cite{NeuralRGBD} and a dense feature grid representation to achieve accurate and accelerated 3D scene reconstruction. We further introduce the \perframe scheme and a \tsdffusion-guided pretraining phase to enhance reconstruction quality, outperforming GO-Surf and Neural RGB-D in quality with minimal impact on optimization time.

\subsection{Camera motion blur handling}
Numerous methods for color image deblurring have been proposed, including those based on convolutional neural networks~\cite{zamir2021multistage} or generative models~\cite{deblurganv2}.
Several studies integrated the deblurring process into NeRF, employing deformable kernel~\cite{dpnerf} or synthesizing blurry images by averaging virtual
sharp color images within learnable exposure time~\cite{BAD_nerf}.

Our approach, distinct from others, centers on correcting distortions in depth frames, which differ markedly from those in color frames~\cite{depth_map_blur}. To tackle depth-specific motion blurs, we utilize \perframe technique, optimizing rendering ray directions through constrained deformable kernels that apply translation and scaling transformations across the entire image planes.

\section{Method}

Our approach is built upon a neural RGB-D surface reconstruction scheme proposed in \cite{NeuralRGBD} and adopts dense feature grid akin to DVGO~\cite{DVGO} for faster optimization. We employ \perframe for correcting camera motion inaccuracies and a three-phase training scheme with \tsdffusion to achieve improved results with reduced computational burden.

\subsection{Hybrid geometry representation}

We employ a dense feature grid to explicitly learn local features, significantly reducing computational complexity and training time compared to using an MLP. We dynamically tailor the feature grid $V^{(feat)}$ for each scene based on scene size--—$L_x$, $L_y$, $L_z$, calculated using depth frames and estimated camera poses:

\begin{equation} \label{eq:dense_feature_grid}
    V^{(feat)}: (N_x \times N_y \times N_z) \mapsto \mathbb{R}^{F},
\end{equation}

\noindent
where $F$ is a fixed hyperparameter for feature vector length and the grid dimensions $N_x$, $N_y$, $N_z$ are set relative to the cell size $gs$, calculated as $\ceil{L_x / gs}, \ceil{L_y / gs}, \ceil{L_z / gs}$. 

For a 3D point $\mathbf{p}$, its feature vector $\mathbf{\phi}_{\mathbf{p}}$ is derived from the trilinear interpolation of the nearest 8 grid vertices $\mathbf{P}_{near}$.

\begin{equation} \label{eq:interpolation}
	\mathbf{\phi}_{\mathbf{p}} = interp[V^{(feat)}(\mathbf{P}_{near})]  \in \mathbb{R}^{F}
\end{equation}

We use shallow MLPs to decode these features into view-dependent color $\radiance$ and truncated signed distance value $\signedDist$ for each 3D point $\mathbf{p}$, as outlined in \cref{eq:shallowmlp_c,eq:shallowmlp_d}. For decoding view-dependent color, positional-encoded ray direction $\mathbf{d}$ and frame-dependent latent embedding vector $\mathbf{\xi}$ are concatenated.

\begin{equation} \label{eq:shallowmlp_c}
	\radiance = MLP^C(\mathbf{\phi}_{\mathbf{p}}, \Lambda(\mathbf{d}), \mathbf{\xi}_i),
\end{equation}
\begin{equation} \label{eq:shallowmlp_d}
	\signedDist = MLP^D(\mathbf{\phi}_{\mathbf{p}}),
\end{equation}
where $i$ is the frame number in the input video and $\Lambda$ is the frequency positional encoding function.

The neural rendering process follows the approach in \cite{NeuralRGBD}. Using a known camera intrinsic matrix, we cast a camera ray $\mathbf{r}$ for each image pixel $(u, v)$ in the normalized image plane along its direction $\mathbf{d}$:

\begin{equation} \label{eq:ray_direction}  
    \mathbf{d}_{u, v} = \left[\begin{array}{@{}c|c@{}} R & t \end{array}\right]
                        \begin{bmatrix} \mathbf{\varrho}_{u,v} \\ 1 \end{bmatrix}
\end{equation}
\begin{equation}
     \mathbf{\varrho}_{u,v} = 
            \begin{bmatrix} 
                {1}/{f_x} & 0 & {-c_x}/{f_x} \\
                0 & {-1}/{f_y} & {c_y}/{f_y} \\
                0 & 0 & -1 
            \end{bmatrix}
            \begin{bmatrix} 
                u \\
                v \\
                1 
            \end{bmatrix}
\end{equation}
\noindent
where $\left[\begin{array}{@{}c|c@{}} R & t \end{array}\right]$ is the estimated camera pose matrix, $f_x, f_y$ are focal lengths and $(c_x, c_y)$ is the principal point.

For a 3D point $\mathbf{p}$, its weight value $\omega_\mathbf{p}$ for rendering the color image is calculated from the signed distance value:
\begin{equation} \label{eq:radiance}
	\omega_\mathbf{p} = \sigma (\frac{\signedDist}{tr}) \cdot \sigma (-\frac{\signedDist}{tr}) ,
\end{equation}
\noindent
where $tr$ is the truncation distance and $\sigma$ is sigmoid function. 

The final rendered color $\widehat{C}$ for a ray $\mathbf{r}$ in the $i^{th}$ frame is computed as the weighted sum of the radiance values of sampled points $\mathbf{p}$ along the ray:
\begin{equation} \label{eq:rendered_color}
	\rayColor = \frac{1}{\sum \omega_\mathbf{p}}\sum\omega_\mathbf{p} \radiance
\end{equation}

\subsection{\Perframe} 
\begin{figure}[t!]
    \centering
    \rotatebox{90}{\makebox[0pt][c]{\small\hspace{5pt} Scene 2}}
    \begin{subfigure}[ht]{0.3\linewidth}
        \includegraphics[width=\textwidth]{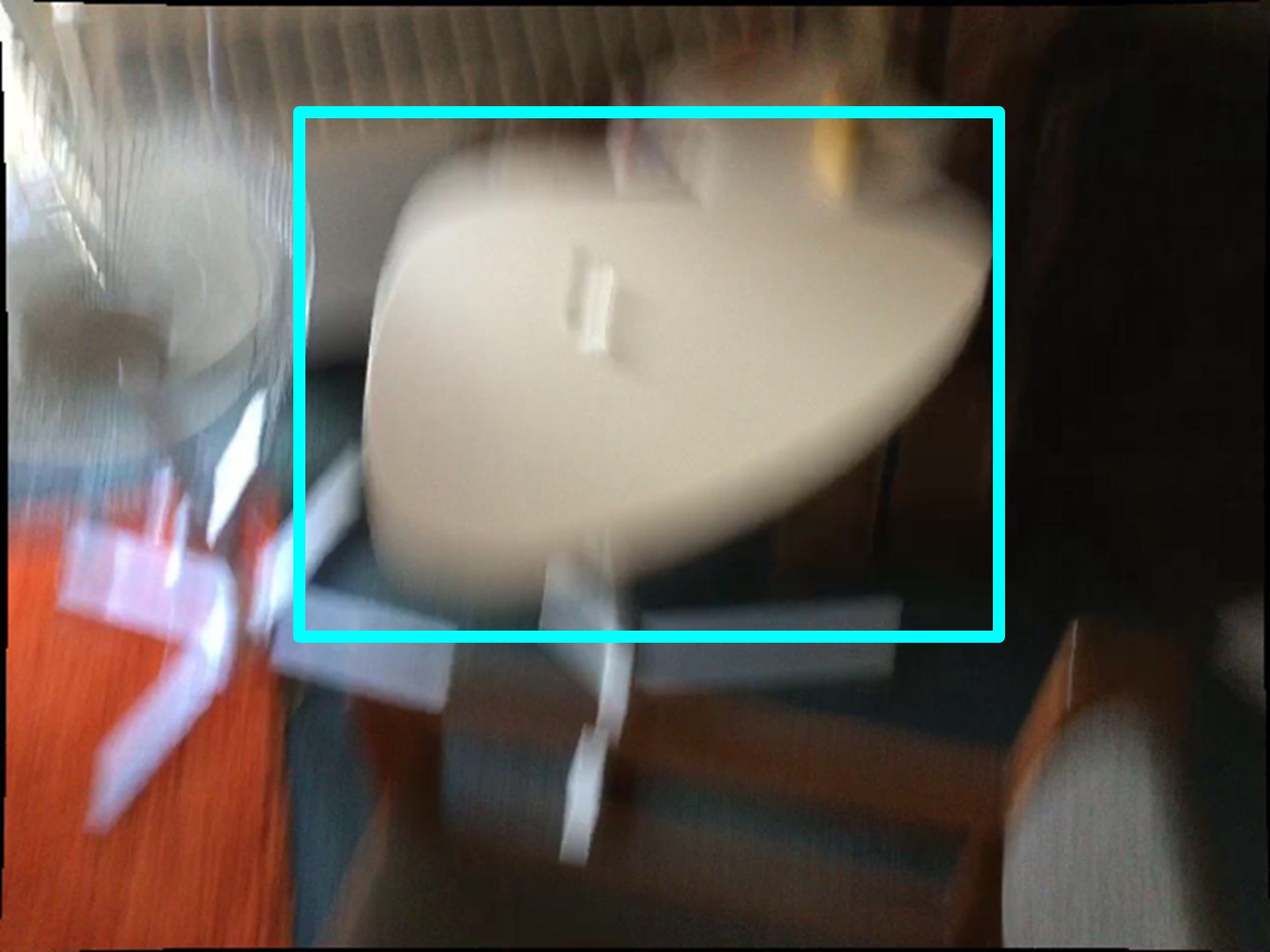}
    \end{subfigure}
    \begin{subfigure}[ht]{0.3\linewidth}
        \includegraphics[width=\textwidth]{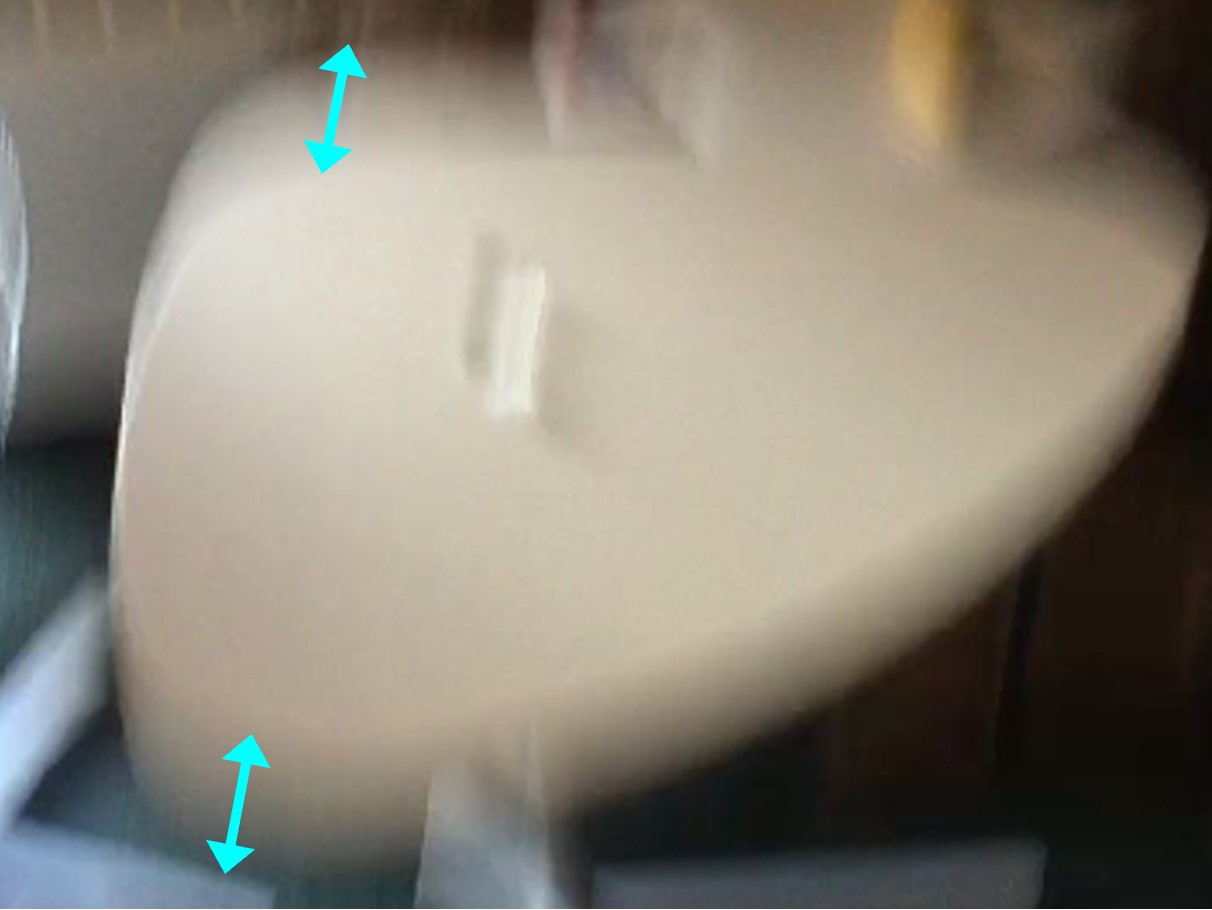}
    \end{subfigure}
    \begin{subfigure}[ht]{0.3\linewidth}
        \includegraphics[width=\textwidth]{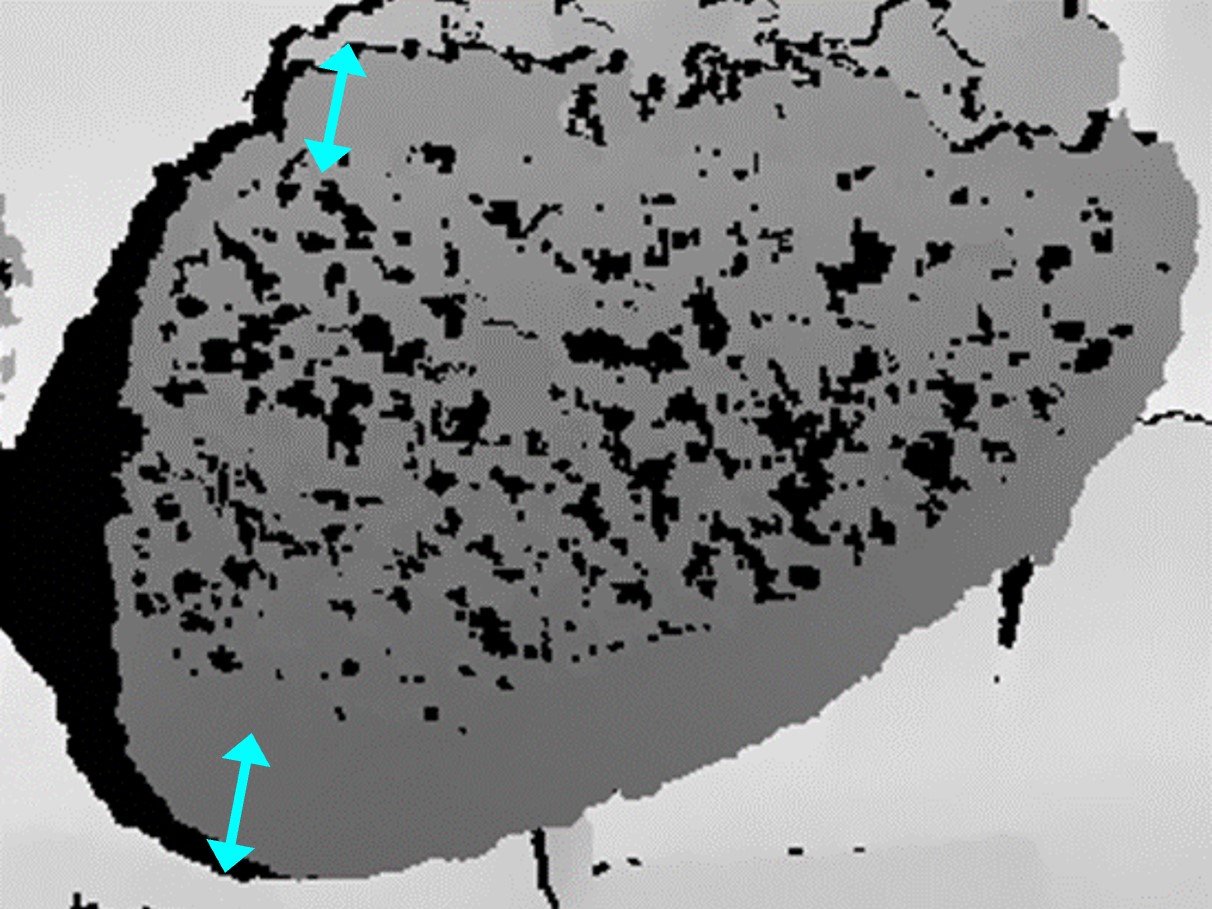}
    \end{subfigure}
    \\
    \rotatebox{90}{\makebox[0pt][c]{\small\hspace{20pt} Scene 11}}
    \begin{subfigure}[ht]{0.3\linewidth}
        \includegraphics[width=\textwidth]{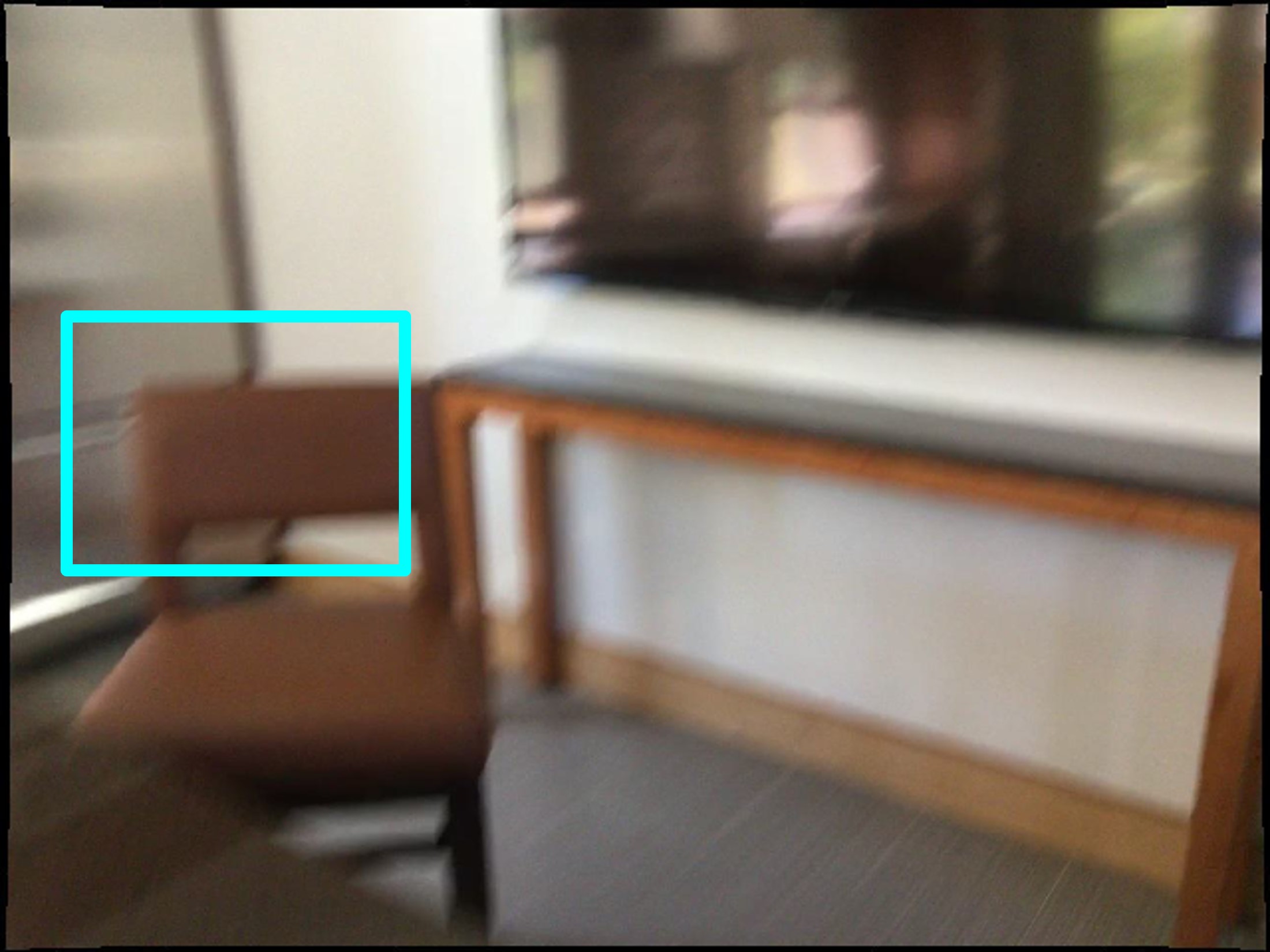}
        \caption{} \label{subfig:RGBBlufFull}
    \end{subfigure}
    \begin{subfigure}[ht]{0.3\linewidth}
        \includegraphics[width=\textwidth]{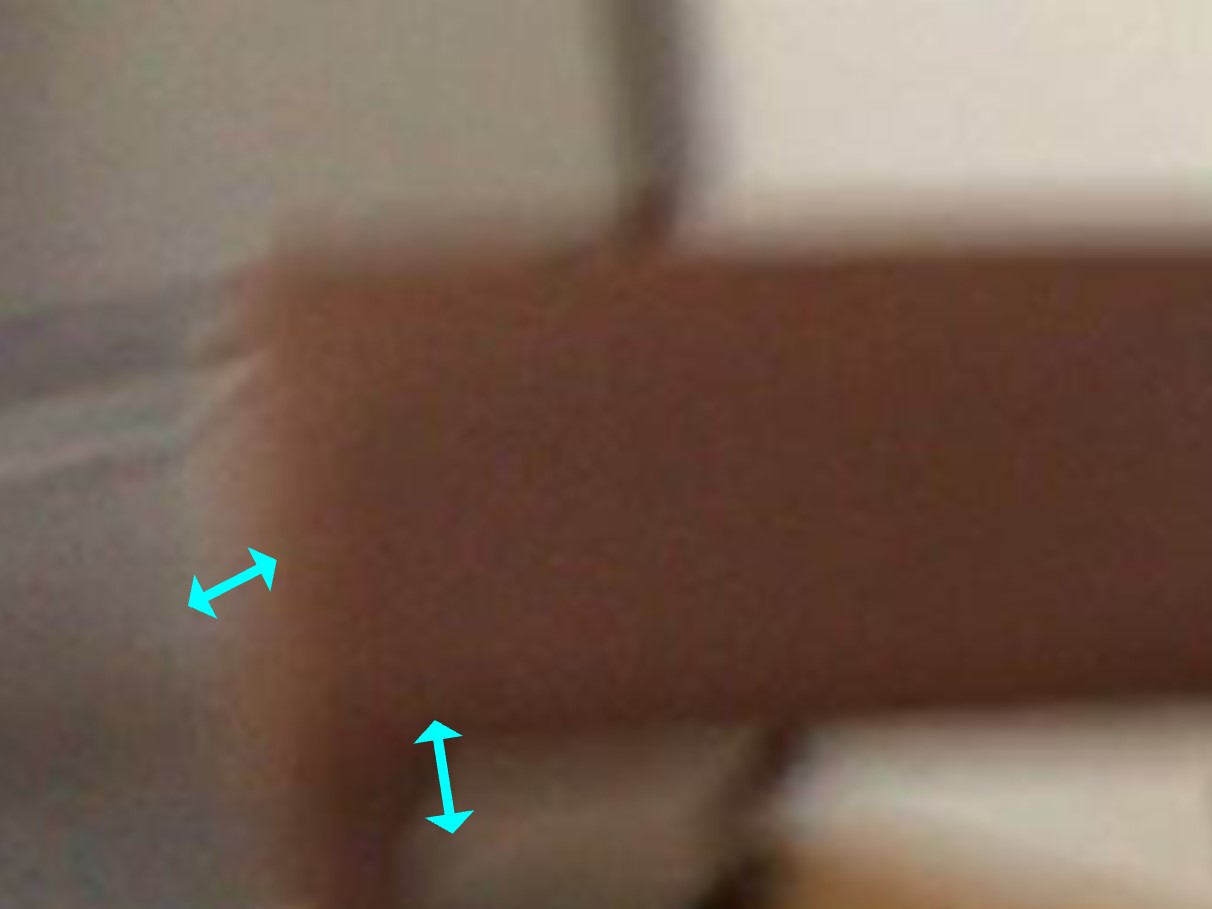}
        \caption{} \label{subfig:RGBBlurCloseup}
    \end{subfigure}
    \begin{subfigure}[ht]{0.3\linewidth}
        \includegraphics[width=\textwidth]{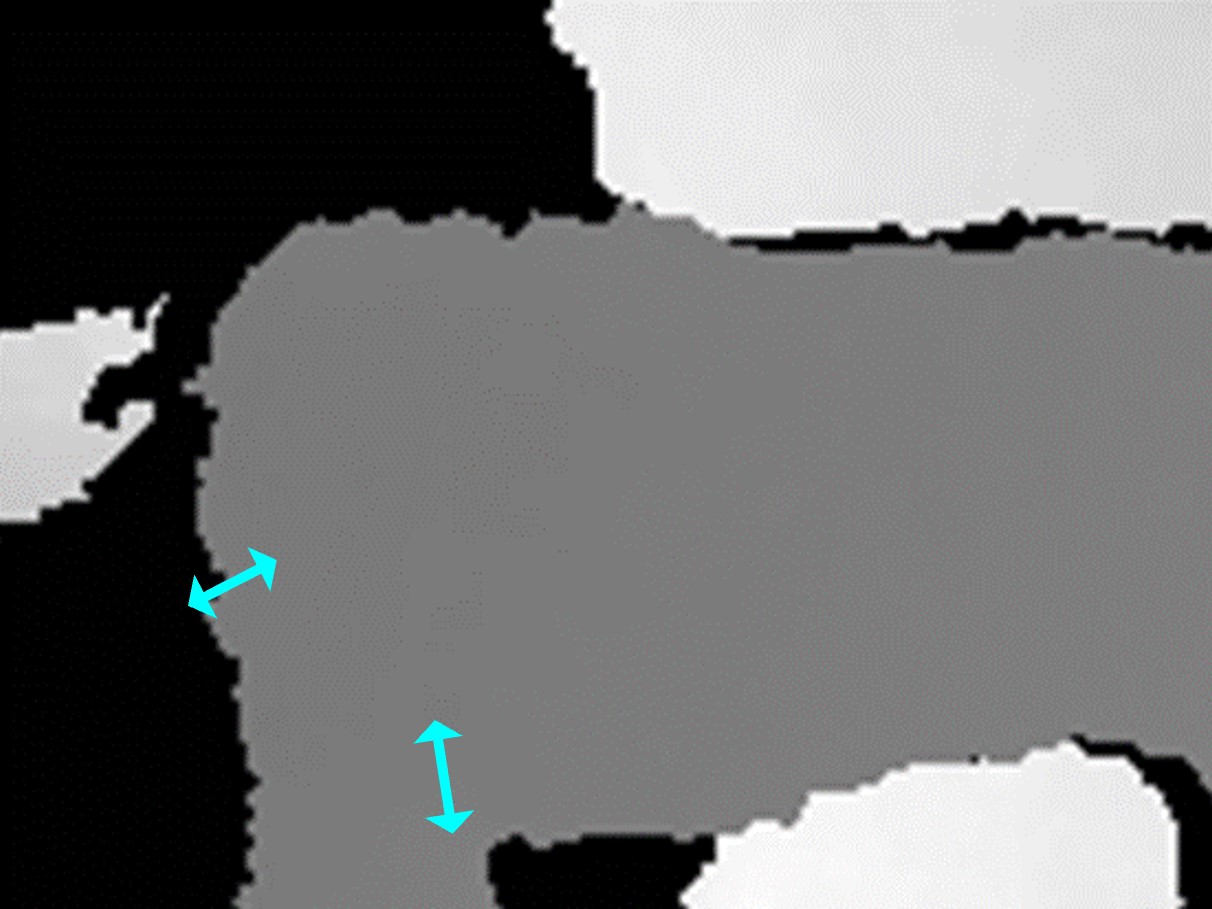}
        \caption{} \label{subfig:depthBlur}
    \end{subfigure}
\caption{Samples from the ScanNet V2~\cite{scannet} dataset demonstrate the negative impact of motion blurs. The RGB frames (a, b) are blurry and distorted. Unlike the color frames, the depth frames (c) contain extended object boundary rather than averaging blur~\cite{depth_map_blur}.}
\label{fig:scannet_misalign}
\end{figure}

In contrast to still images, video frames are susceptible to camera motion, resulting in motion blurs as depicted in \cref{fig:scannet_misalign}. While motion blurs in color frames are typically modeled as averaging pixels over exposure time, motion blurs in depth frames act more like a min-filter, taking the minimum value during exposure time. This phenomenon can cause the boundaries to extend beyond actual object boundaries~\cite{depth_map_blur}.
Our method is designed to handle motion blur in depth frames by correcting the camera intrinsic matrix, which effectively changes the scale and translation of the projected image plane.

Before handling this \perframe issue, we first adopt the image-plane deformation field and pose optimization techniques from \cite{NeuralRGBD} to correct the camera-pose errors and the potential global errors from the intrinsic parameters as well as the camera lens distortion. We modify the image-plane deformation field to use a two-layered shallow MLP to reduce the training time. 

Our \perframe scheme is performed after the image-plane deformation field has been applied to correct the depth-specific motion blurs.
Specifically, we introduce four parameters per frame, two for scaling and the other two for translation purposes.
Scaling and translation are applied to the normalized image coordinate before it is transformed into the world coordinate:
\begin{gather} \label{eq:align_optim}
	\mathbf{\widehat{\varrho}}_{u,v}^{\,i} = \mathbf{s}^{i} \cdot (\mathbf{\varrho}_{u,v} + \mathbf{\tau}^{i}) \\
        \label{eq:align_param}
	\mathbf{s}^i = \begin{bmatrix} 
                s_x^i & 0 & 0 \\ 0 & s_y^i & 0 \\ 0 & 0 & 1 
                \end{bmatrix},\,\,
    \mathbf{\tau}^i = \begin{bmatrix} \tau_x^i \\ \tau_y^i \\ 0 \end{bmatrix}
\end{gather}

\noindent
where $\mathbf{s}^i$ and $\mathbf{\tau}^i$ are trainable parameters for $i^{th}$ frame and optimized during training. Accordingly, the refined casting direction $\widehat{\mathbf{d}}$ is calculated from $\mathbf{\widehat{\varrho}}$:
\begin{equation} \label{eq:refined_direction}
    \widehat{\mathbf{d}}_{u, v}^{\,i} = 
        \left[\begin{array}{@{}c|c@{}} R & t \end{array}\right]
        \begin{bmatrix} \mathbf{\widehat{\varrho}}_{u,v}^{\,i} \\ 1 \end{bmatrix}
\end{equation}
\cref{eq:refined_direction} replaces \cref{eq:ray_direction} for generating rays and sampling points. These parameters can be interpreted as correcting the principal point and focal lengths of the intrinsic matrix.

The \ipdf and \perframe schemes both aim to refine ray directions for accuracy. \ipdf globally refines across all frames, whereas \perframe targets frame-specific fluctuations not addressed by \ipdf.

\subsection{Optimization}
\infusionsurf is optimized through a three-phase training process as depicted in~\cref{fig:teaser}.
In the first phase, \infusionsurf learns a geometric prior using the classical real-time algorithm, \tsdffusion~\cite{SDF}. 
Specifically, \infusionsurf builds a dense grid as in \cref{eq:dense_feature_grid} with $F$=1 and runs \tsdffusion with depth frames of the identical scene.
During optimization, we randomly sample grid cells, query the center points for signed distance values, and strive to minimize the mean squared error against \tsdffusion's corresponding values.
Only parameters of $V^{(feat)}$ and $MLP^D$ are trained during this phase.

The optimization phase enables direct learning of signed distance values, bypassing the time-consuming rendering process and eventually leading to a substantial acceleration of the training phase.

In the second and third phases of \infusionsurf, we adopt progressive learning similar to ~\cite{DVGO} to sequentially refine low- and high-frequency details. All parameters, including those not involved in the first phase, undergo optimized in these stages.

During phase two, \infusionsurf randomly samples ray batches $r_b$ and points $S_c$ every 15.625mm along the rays. Our loss function, similar to \cite{NeuralRGBD}, uses estimated signed distance values and rendered colors (\cref{eq:shallowmlp_d,eq:rendered_color}):

\begin{equation} \label{eq:loss_all}
	\mathcal{L} = \lambda_{fs}\mathcal{L}_{fs} + \lambda_{sdf}\mathcal{L}_{sdf} + \lambda_{rgb}\mathcal{L}_{rgb} + \lambda_{reg}\mathcal{L}_{reg} 
 \end{equation}

$\mathcal{L}_{fs}$ and $\mathcal{L}_{sdf}$ represent loss components for the points outside ($S_{fs}$) and within ($S_{sdf}$) the truncated area respectively:

\begin{equation} \label{eq:loss_fs}
    \mathcal{L}_{fs} = \frac{1}{|r_b|}\sum_{r \in r_b}\frac{1}{|S_{fs}|}\sum_{\mathbf{p} \in S_{fs}}(\widehat{D}_{\mathbf{p}} - tr)^2,
\end{equation}

\begin{equation} \label{eq:loss_sdf}
    \mathcal{L}_{sdf} = \frac{1}{|r_b|}\sum_{r \in r_b}\frac{1}{|S_{sdf}|}\sum_{\mathbf{p} \in S_{sdf}}(\widehat{D}_{\mathbf{p}} - D_\mathbf{r}^i)^2
\end{equation}
\noindent
where $D_\mathbf{r}^i$ is the signed distance value observed in the depth frame.

$\mathcal{L}_{rgb}$ measures the difference between the rendered color and the observed color of the corresponding pixels:

\begin{equation} \label{eq:loss_rgb}
    \mathcal{L}_{rgb} = \frac{1}{|r_b|}\sum_{r \in r_b}(\rayColor - C_{u,v}^{\,i})^2
\end{equation}

The term $\mathcal{L}_{reg}$ denotes the L2 regularization for the frame-dependent latent embedding vector, \perframe parameters, and image-plane deformation parameters. Notably, the scaling parameters in \perframe undergo regularization centered around 1.

In the third phase, \infusionsurf focuses on fine details by dividing the dense feature grid into higher resolutions, halving the grid cell size $gs$. Additional points $S_f$ are sampled around surfaces identified in $S_c$. This phase employs the same loss function (\cref{eq:loss_all}), utilizing both $S_c$ and $S_f$ for training.

\subsection{Implementation details}

Our network, built with PyTorch, was optimized using ADAM \cite{adam_optim} with initial settings of a $5 \times 10^{-4}$ learning rate (decaying exponentially to a tenth every 250K iterations), 0.9 beta1, and 0.999 beta2. We set the feature grid with $F$=12 and $gs$=10cm, initializing weights at 0, and truncated signed distance values at $tr$=5cm. Both color and signed distance MLPs had 2 hidden layers with 128 nodes, while the image-plane deformation MLP had 2 layers with 64 nodes each. \Perframe parameters started at 1 for scaling and 0 for translation. Loss term weights were $\lambda_{fs}$=10, $\lambda_{sdf}$=$6 \times 10^3$, $\lambda_{rgb}$=0.5, and $\lambda_{reg}$=0.1. We ran 3K, 7K, and 65K iterations for each training phase, respectively. We used a CUDA implementation of the \tsdffusion algorithm that can process, on average, 231.7 frames per second on Tesla V100 GPU \cite{3DMatch}.
\section{Experiments}

We demonstrate comparative studies of \infusionsurf against prior work as well as an ablation study of the proposed framework components to show the impact of \perframe and \tsdffusion-guided training phase. Please refer to Appendix section for additional experiments and results.

For the evaluation of each study, we extracted the truncated signed distance values in $1\text{cm}^3$ resolution and ran MarchingCubes~\cite{marching_cube} algorithm to get the triangular meshes.

\subsection{Datasets}
We used ScanNet V2 \cite{scannet} as a real-scene dataset during the qualitative study.
The dataset used the rolling shutter method during image capture, resulting in motion blur, distortions, and noisy depth values, including holes and missing objects.

To compare quantitative results against the baseline methods, we used 10 synthetic scenes from \cite{NeuralRGBD}. They used indoor 3D models to photo-realistically render color and depth frames with ground truth camera trajectories.
Depth frames were simulated with Kinect-like noises including holes and quantization noises.

\subsection{Baselines}
We compare our results with NeRF-style RGB-D 3D surface reconstruction methods, Neural RGB-D \cite{NeuralRGBD} and GO-Surf \cite{GOSurf}.
Specifically, we report our results at 20K and 75K iterations to respectively compare against GO-Surf and Neural RGB-D to show that \infusionsurf outperforms in terms of both efficiency and performance.
We trained GO-Surf and Neural RGB-D with the hyperparameters recommended by the respective papers on a Tesla V100 GPU.

\subsection{Results and discussion}
\subsubsection{Qualitative results} 
\begin{figure*}[ht!]
    \rotatebox{90}{\makebox[0pt][c]{\small \hspace{0pt} Scene 2}}
    \centering
    \begin{subfigure}[ht]{0.23\linewidth}
        \includegraphics[width=\linewidth]{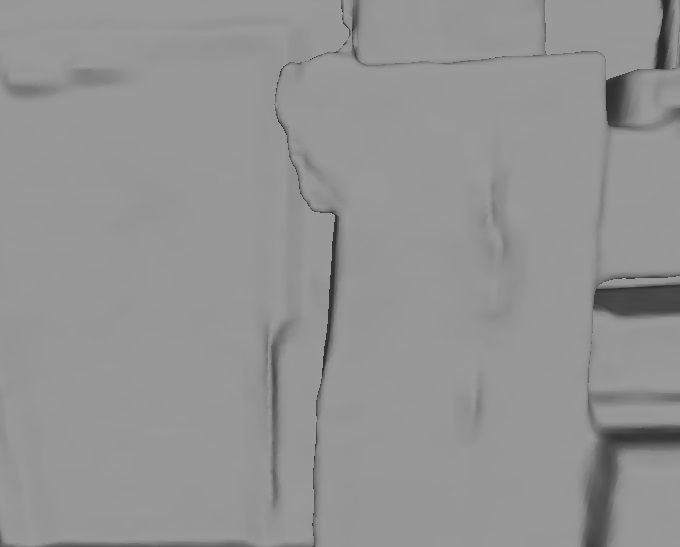}
        \centering
    \end{subfigure}
    \begin{subfigure}[ht]{0.23\linewidth}
        \includegraphics[width=\linewidth]{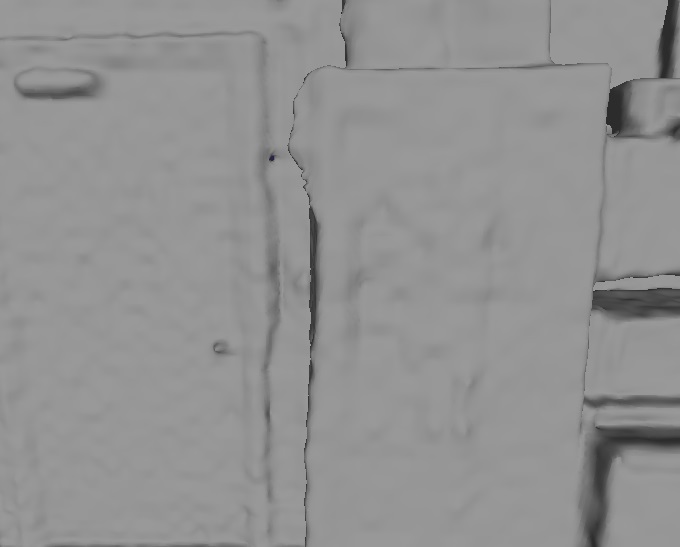}
        \centering
    \end{subfigure}
    \begin{subfigure}[ht]{0.23\linewidth}
        \includegraphics[width=\linewidth]{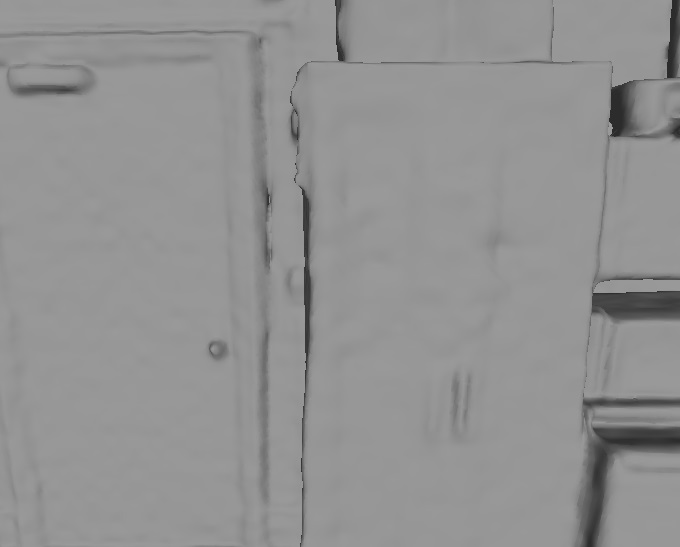}
        \centering
    \end{subfigure}
    \begin{subfigure}[ht]{0.23\linewidth}
        \includegraphics[width=\linewidth]{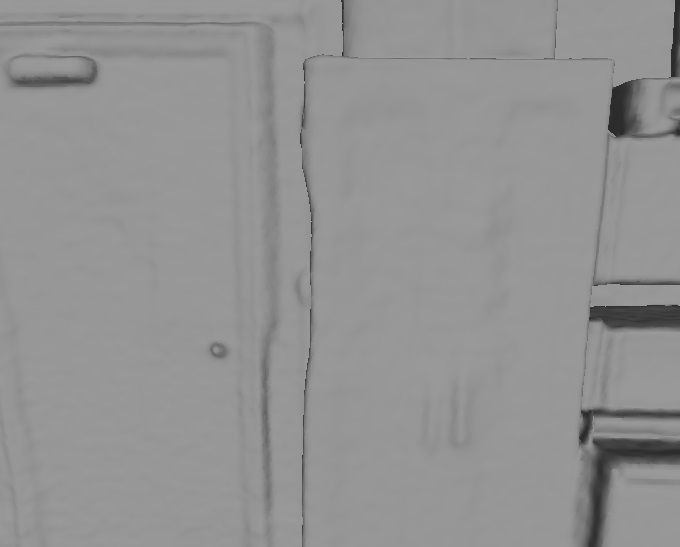}
        \centering
    \end{subfigure}
    \\
    \rotatebox{90}{\makebox[0pt][c]{\small Scene 5}}
    \begin{subfigure}[ht]{0.23\linewidth}
        \includegraphics[width=\linewidth]{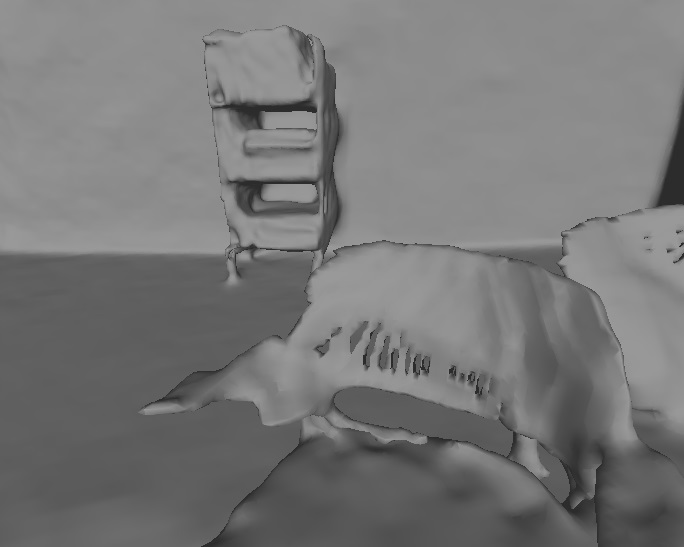}
        \centering
    \end{subfigure}
    \begin{subfigure}[ht]{0.23\linewidth}
        \includegraphics[width=\linewidth]{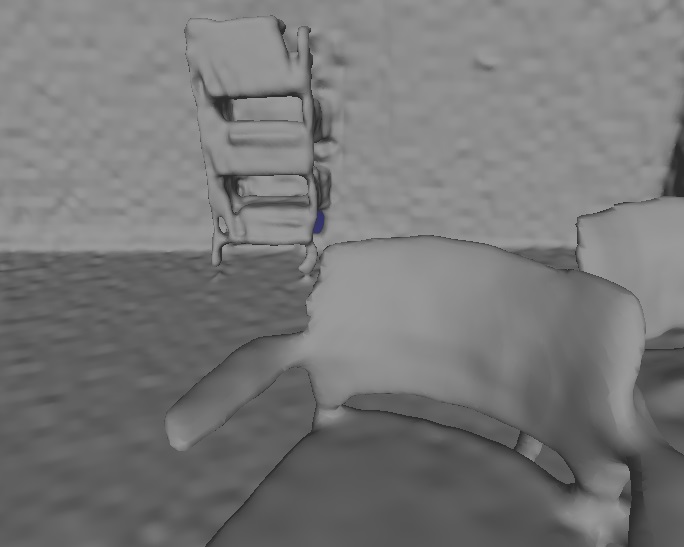}
        \centering
    \end{subfigure}
    \begin{subfigure}[ht]{0.23\linewidth}
        \includegraphics[width=\linewidth]{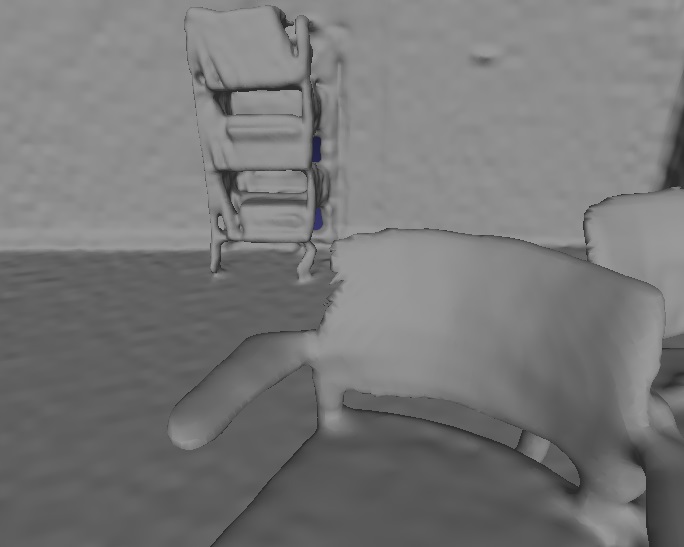}
        \centering
    \end{subfigure}
    \begin{subfigure}[ht]{0.23\linewidth}
        \includegraphics[width=\linewidth]{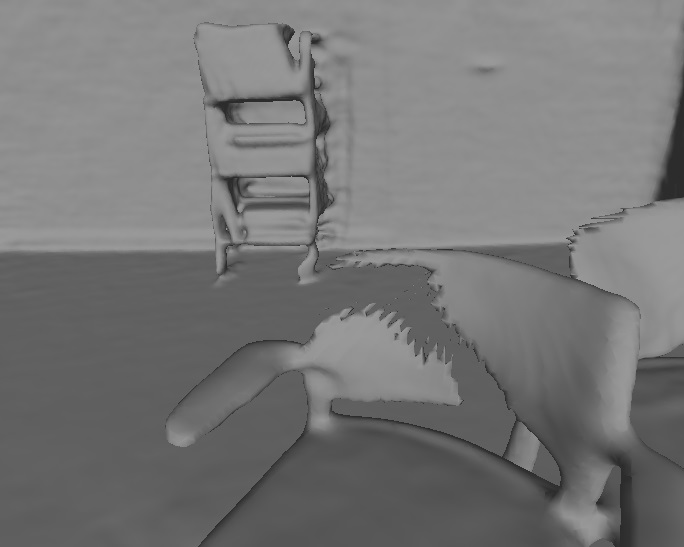}
        \centering
    \end{subfigure}
    \\
    \rotatebox{90}{\makebox[0.01\linewidth][c]{\small Scene 12}}
    \begin{subfigure}[ht]{0.23\linewidth}
        \includegraphics[width=\linewidth]{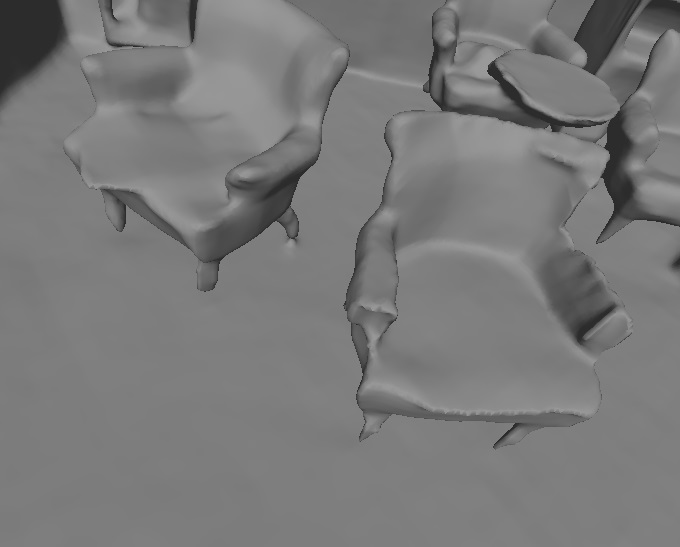}
        \centering
    \end{subfigure}
    \begin{subfigure}[ht]{0.23\linewidth}
        \includegraphics[width=\linewidth]{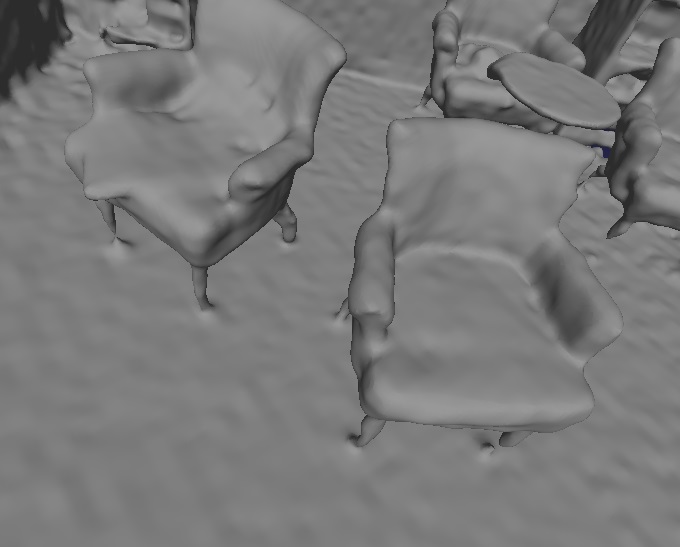}
        \centering
    \end{subfigure}
    \begin{subfigure}[ht]{0.23\linewidth}
        \includegraphics[width=\linewidth]{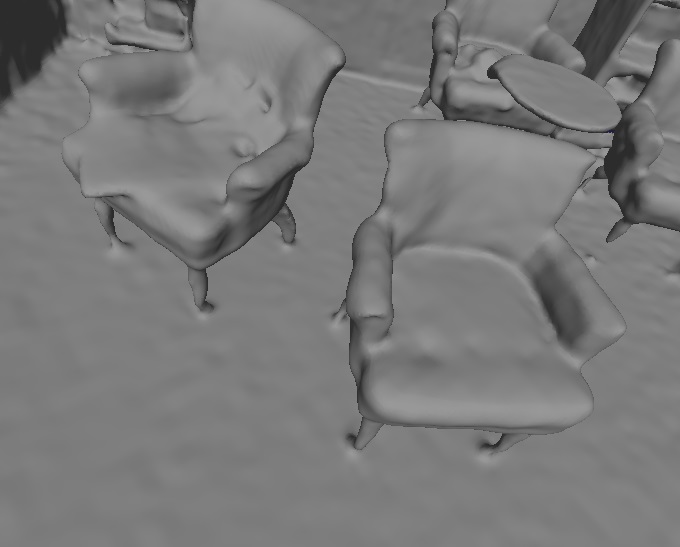}
        \centering
    \end{subfigure}
    \begin{subfigure}[ht]{0.23\linewidth}
        \includegraphics[width=\linewidth]{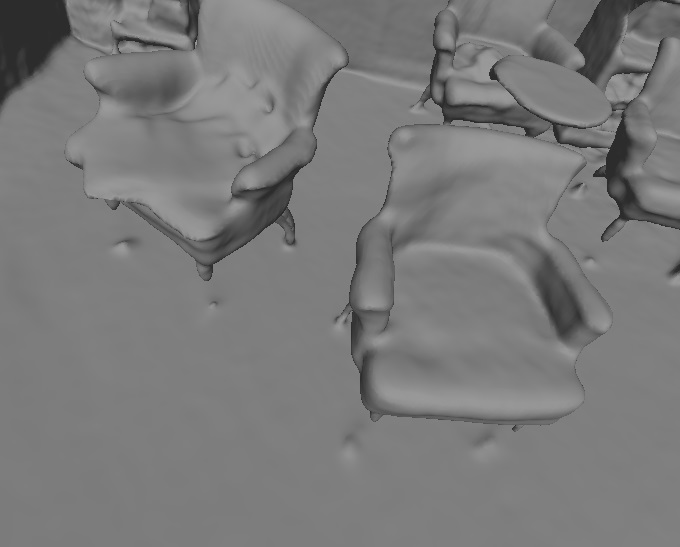}
        \centering
    \end{subfigure}
    \\
    \rotatebox{90}{\makebox[0.01\linewidth][c]{\small \hspace{30pt}Scene 50}}
    \begin{subfigure}[ht]{0.23\linewidth}
        \includegraphics[width=\linewidth]{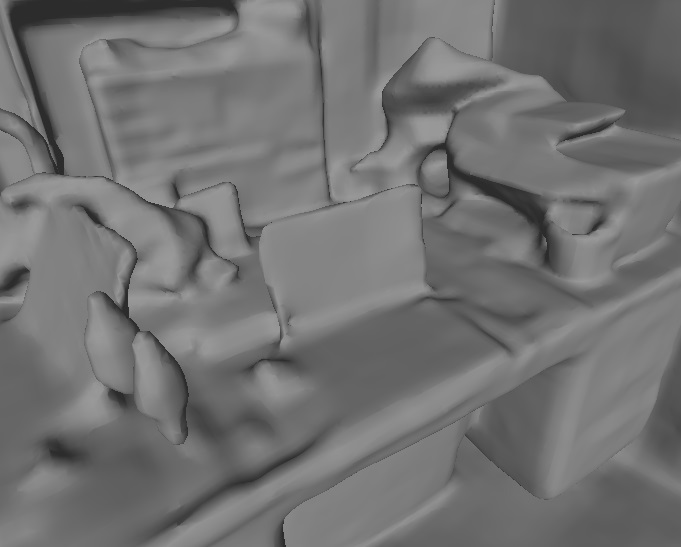}
        \centering
        \caption*{Train time: 19m15s - 22m43s}
        \caption{GO-Surf} \label{caption:GoSurf}
        \label{subfig:go_surf_qual}
    \end{subfigure}
    \begin{subfigure}[ht]{0.23\linewidth}
        \includegraphics[width=\linewidth]{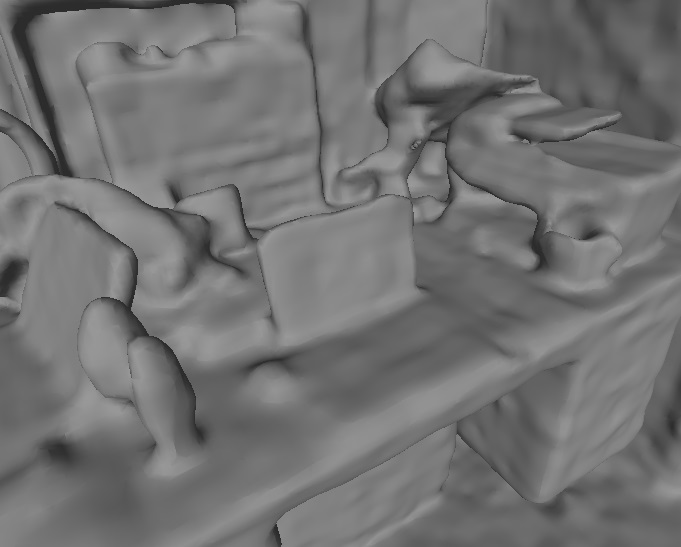}
        \centering
        \caption*{11m08s - 14m30s}
        \caption{\textbf{Ours: 20K iterations}} \label{caption:OursShort}
        \label{subfig:fast_surf_fast_qual}
    \end{subfigure}
    \begin{subfigure}[ht]{0.23\linewidth}
        \includegraphics[width=\linewidth]{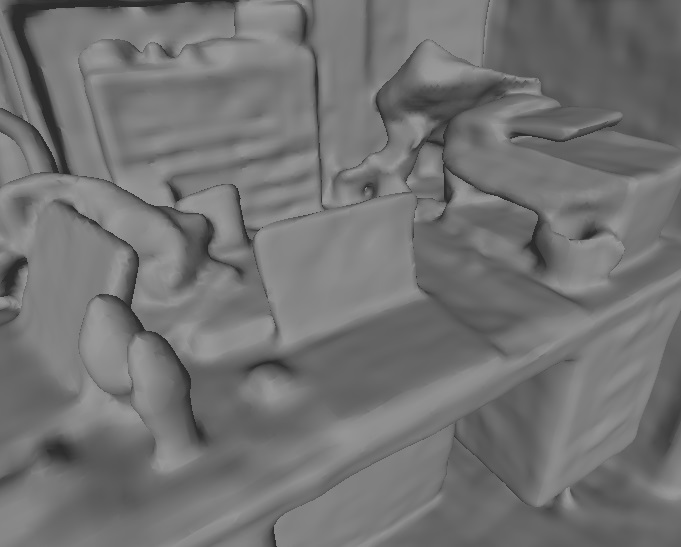}
        \centering
        \caption*{54m35s - 73m30s}
        \caption{\textbf{Ours: 75K iterations}} \label{caption:OursLong}
        \label{subfig:fast_surf_slow_qual}
    \end{subfigure}
    \begin{subfigure}[ht]{0.23\linewidth}
        \includegraphics[width=\linewidth]{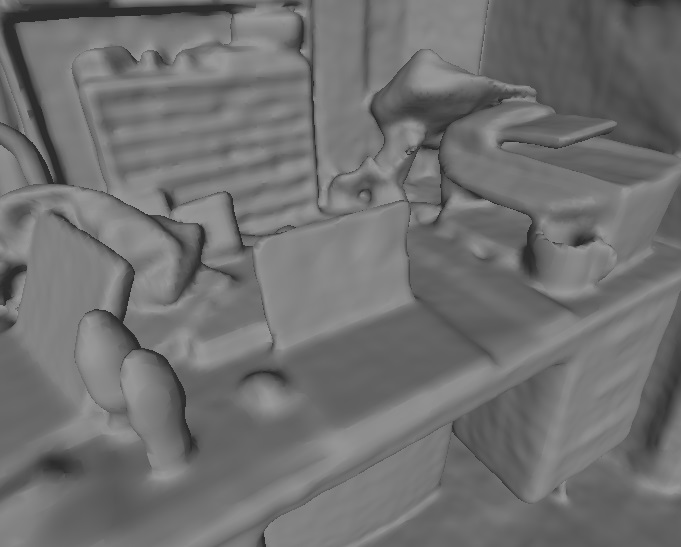}
        \centering
        \caption*{452m10s - 546m50s}
        \caption{Neural RGB-D} \label{caption:NeuralRGBD}
        \label{subfig:neural_rgbd_qual}
    \end{subfigure}
    \caption{
    We compare our method with GO-Surf~\cite{GOSurf} and Neural RGB-D~\cite{NeuralRGBD} at different points in time. The comparison was conducted using scenes 2, 5, 12, and 50 from ScanNet V2~\cite{scannet}. When trained for a shorter amount of time, \infusionsurf-20K (b) recovers high-frequency details overlooked by GO-Surf (a) and generates much less erroneous surfaces. Given a longer training time, \infusionsurf-75K (c) achieves greater quality while recovering a number of geometries missing from Neural RGB-D (d).
    }
    \label{fig:qualitative_result}
\end{figure*}

The qualitative results are illustrated in \cref{fig:qualitative_result}.
For the comparison with GO-Surf, our reconstruction results after 20K iterations is shown.
As depicted in \cref{fig:qualitative_result}, \infusionsurf exhibits finer details and fewer erroneous surfaces throughout the scenes, even with a shorter training time.

After 75K iterations, our reconstruction qualities showed better results than Neural RGB-D, recovering the structures Neural RGB-D missed in some cases.
Compared to Neural RGB-D, our training speed was 7.3 to 8.7 times faster. 

\subsubsection{Quantitative results} We compared our method with baselines in terms of Chamfer $\ell_1$ distance (C-$\ell_1$), intersection-over-union (IoU), normal consistency (NC), and F-score.
In order to compute C-$\ell_1$, NC, and F-score, we sampled point clouds from the output meshes in 1$\text{cm}^2$ resolution. We used a threshold of 5cm for F-score.
The evaluation meshes were voxelized with an edge length of 10cm to compute the IoU.
As shown in \cref{tab:quantitative_result}, our result after 20K iterations outperforms GO-Surf in C-$\ell_1$, IoU, and F-score with less training time (20\% faster on average).
In the comparison with Neural RGB-D, our method also showed superior C-$\ell_1$, IoU, and F-score. At the same time, it took 96 minutes on average to train 75K iterations, which was 7 times faster than what Neural RGB-D took.
While \infusionsurf showed outstanding performance on the three major measures, it was less effective on the normal consistency, indicating that its results contained relatively uneven surfaces.
The quantitative metrics imply that our method is best suited for quickly reconstructing complex geometries, rather than simple flat surfaces.

\begin{figure}[ht!]
    \centering
    \rotatebox{90}{\makebox[0pt][c]{\small\hspace{20pt} Scene 5}}
    \begin{subfigure}[ht]{0.31\linewidth}
        \includegraphics[width=\linewidth]{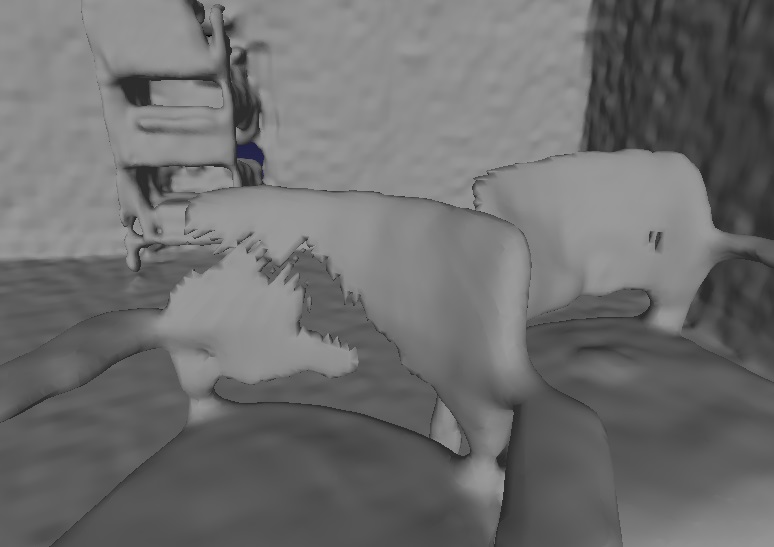}
        \caption*{7s (TSDF) / 14m47s}
        \centering
    \end{subfigure}
    \begin{subfigure}[ht]{0.31\linewidth}
        \includegraphics[width=\linewidth]{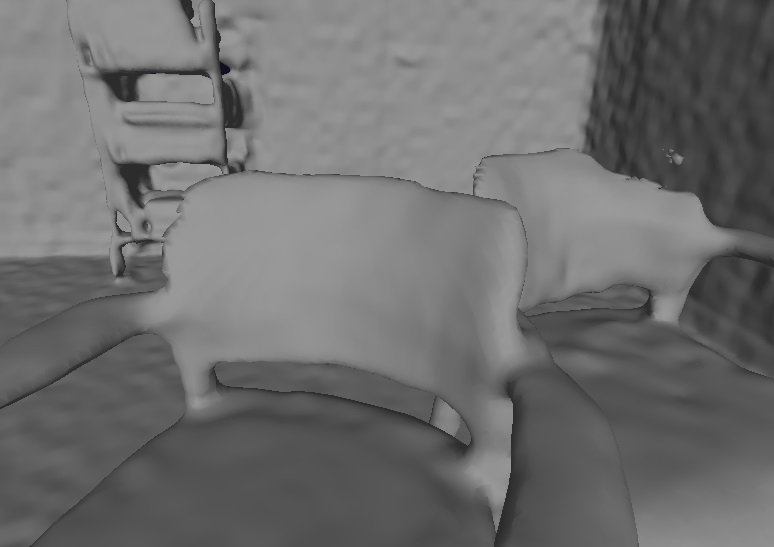}
        \caption*{17m06s}
        \centering
    \end{subfigure}
    \begin{subfigure}[ht]{0.31\linewidth}
        \includegraphics[width=\linewidth]{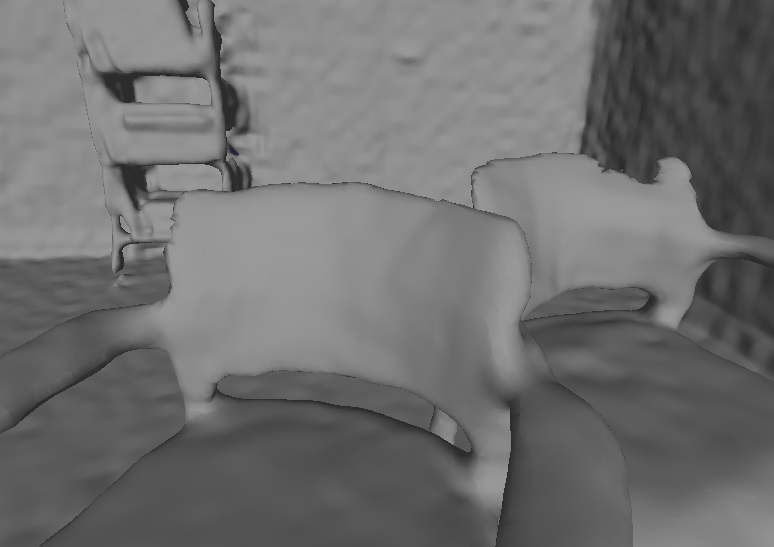}
        \caption*{7s / 15m16s}
        \centering
    \end{subfigure}
    \\
    \rotatebox{90}{\makebox[0pt][c]{\small\hspace{20pt} Scene 0}}
    \begin{subfigure}[ht]{0.31\linewidth}
        \includegraphics[width=\linewidth]{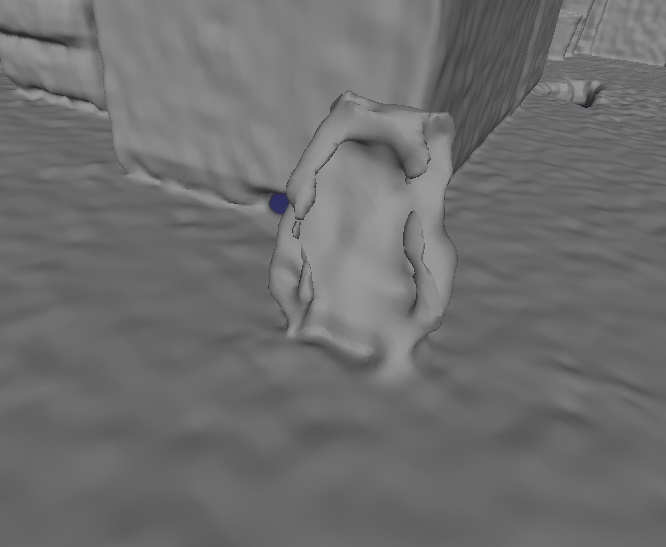}
        \caption*{15s / 34m05s}
        \centering
    \end{subfigure}
    \begin{subfigure}[ht]{0.31\linewidth}
        \includegraphics[width=\linewidth]{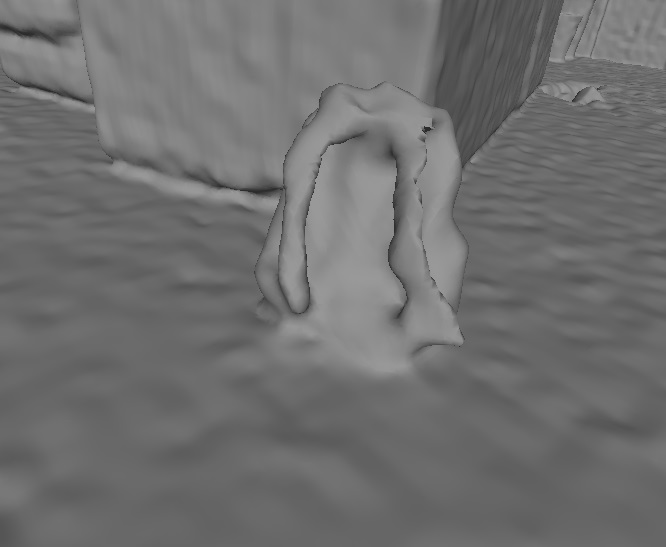}
        \caption*{37m55s}
        \centering
    \end{subfigure}
    \begin{subfigure}[ht]{0.31\linewidth}
        \includegraphics[width=\linewidth]{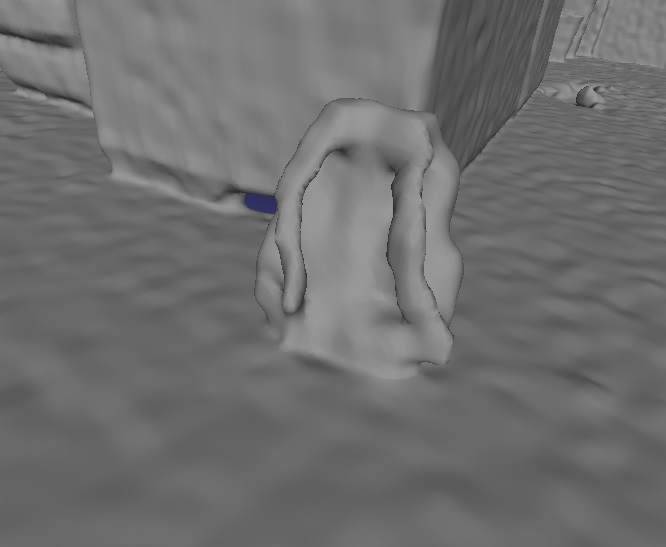}
        \caption*{15s / 34m51s}
        \centering
    \end{subfigure}
    \\
    \rotatebox{90}{\makebox[0pt][c]{\small\hspace{20pt} Scene 2}}
    \begin{subfigure}[ht]{0.31\linewidth}
        \includegraphics[width=\linewidth]{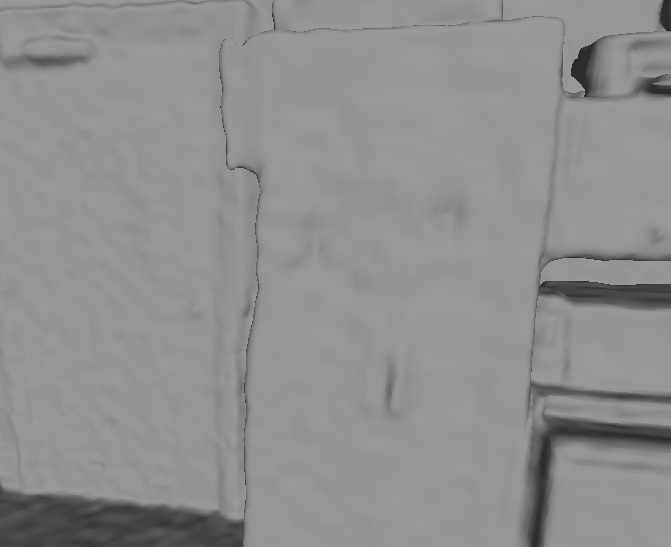}
        \caption*{9s / 17m11s}
        \centering
    \end{subfigure}
    \begin{subfigure}[ht]{0.31\linewidth}
        \includegraphics[width=\linewidth]{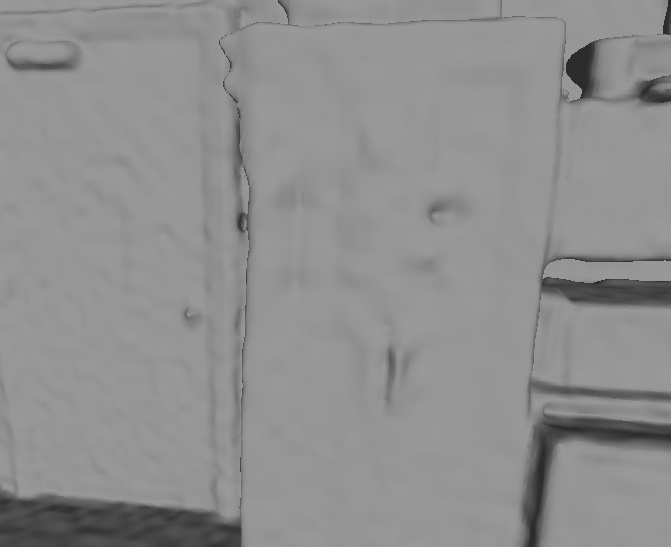}
        \caption*{19m21s}
        \centering
    \end{subfigure}
    \begin{subfigure}[ht]{0.31\linewidth}
        \includegraphics[width=\linewidth]{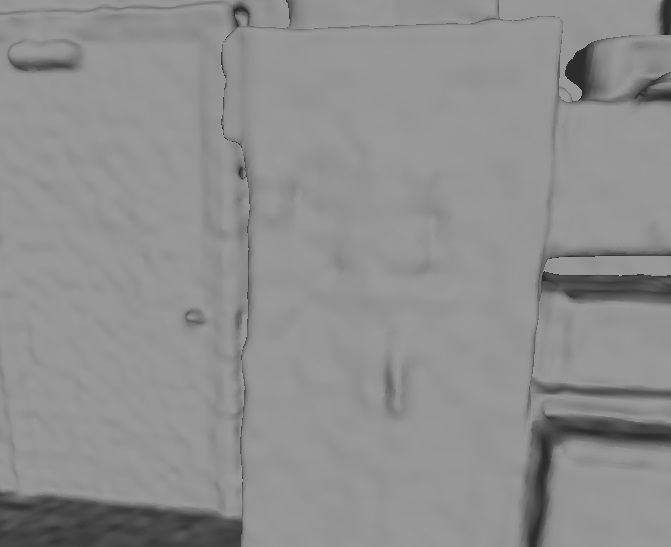}
        \caption*{9s / 17m39s}
        \centering
    \end{subfigure}
    \\
    \rotatebox{90}{\makebox[0pt][c]{\small\hspace{36pt} Scene 50}}
    \begin{subfigure}[ht]{0.31\linewidth}
        \includegraphics[width=\linewidth]{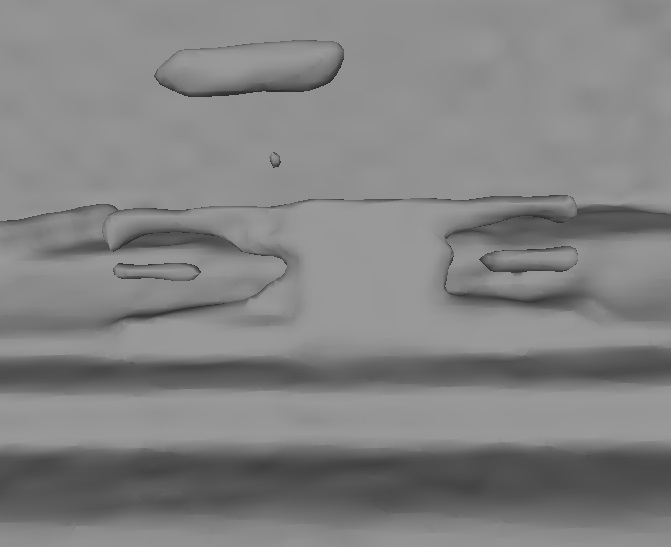}
        \caption*{9s / 16m42s}
        \caption{Ours w/o PFIR}
        \centering
        \label{subfig:ablation_no_align}
    \end{subfigure}
    \begin{subfigure}[ht]{0.31\linewidth}
        \includegraphics[width=\linewidth]{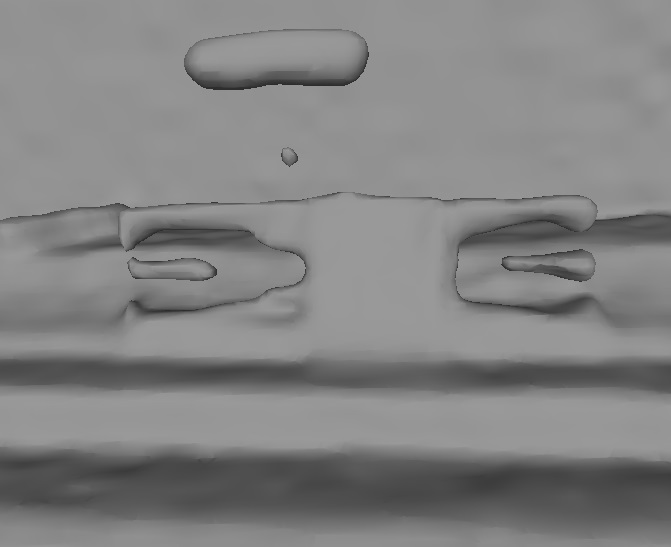}
        \caption*{19m24s}
        \caption{Ours w/o TSDF}
        \centering
        \label{subfig:ablation_no_init}
    \end{subfigure}
    \begin{subfigure}[ht]{0.31\linewidth}
        \includegraphics[width=\linewidth]{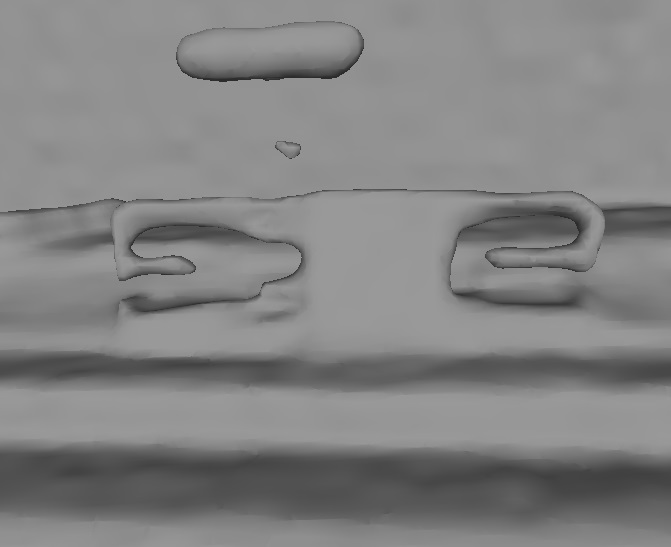}
        \caption*{9s / 16m66s}
        \caption{Ours full}
        \centering
        \label{subfig:ablation_full}
    \end{subfigure}
    \caption{Ablation study. (a) Ours without \perframe (PFIR). (b) Ours without \tsdffusion prior learning in the first phase of training (TSDF). (c) Ours with both methods applied. Timestamps below the subfigures represent the \tsdffusion prior learning time (if applicable) and the total training time.}
    \label{fig:ablation_study}
\end{figure}

\begin{table*}[ht!]
\centering
\begin{tabular}{lccccc}
\hline
\multicolumn{1}{c}{\textbf{Method}} & \textbf{C-$\ell_1$} $\downarrow$         & \textbf{IoU} $\uparrow$            & \textbf{NC} $\uparrow$             & \textbf{F-Score} $\uparrow$        & \textbf{Time} \\ \hline
GO-Surf                & 0.042    & 0.723          & \textbf{0.922} & 0.918          &  22m57s                 \\
Neural RGB-D           & 0.052          & \underline{0.757}          & \textbf{0.922} & \underline{0.938} & 669m59s                \\ \hline
Ours (20K)     & \textbf{0.038}    & 0.737          & 0.904          & 0.929          & 
  18m24s                 \\
Ours (20K, w/o TSDF)       & 0.042    & 0.750         & 0.902          & 0.926          &  21m05s                 \\
Ours (75K)      & \underline{0.041} & \textbf{0.768} & \underline{ 0.913}    & \textbf{0.939}    &  95m57s                 \\ \hline

\end{tabular}

\caption{Quantitative results on the synthetic scene dataset. \infusionsurf shows better C-$\ell_1$, IoU, and F-score than GO-Surf~\cite{GOSurf}  when trained for a shorter amount of time. After training for more iterations, it achieves better performances than Neural RGB-D~\cite{NeuralRGBD} with significantly less training time. The performance 20K iterations without \tsdffusion (TSDF) implies that our \tsdffusion-guided training phase improves both reconstruction qualities and training time. }
\label{tab:quantitative_result}
\end{table*}

\subsubsection{Ablation study}
\label{sec:ablation}
The ablation study shows the effects of \perframe and \tsdffusion-guided training phase.
In order to qualitatively evaluate, we compared results at the same iteration points without the \perframe or the first phase of training---\tsdffusion prior learning (\cref{fig:ablation_study}).
To quantitatively evaluate the effectiveness of the \tsdffusion, we conducted a quantitative study with and without the \tsdffusion-guided training phase (\cref{tab:quantitative_result}).

As shown in \cref{fig:ablation_study}, our \perframe method fixes the erroneous parts of the objects. 
Our method adds only four additional parameters per frame, which can be optimized in the early stages of training. This fixes reconstruction errors in small iterations, increasing the training time by just 2.7\% on average.

As illustrated in \cref{fig:ablation_study}, the \tsdffusion prior learning phase consistently improves object reconstruction details across all scenes. Moreover, the lack of \tsdffusion prior learning results in several minutes of additional training time for the same iteration. This consistent pattern is further supported by the results presented in \cref{tab:quantitative_result}, where our approach excels in C-$\ell_1$, NC, and F-score metrics with improved training time.
The first phase of our training process, leveraging prior knowledge, is extremely fast because it directly learns the signed distance values of 3D coordinates instead of requiring the time-consuming rendering process, albeit with some inaccuracies.
The subsequent training phases can dedicate their efforts to refining intricate details within the scene, ultimately improving reconstruction quality while requiring less time to converge.

\subsection{Limitation}

A main limitation of our approach lies in its reliance on simple transformation matrices to correct motion blurs from camera movements, effectively addressing uniform frame distortions but not local blurs like object motion or rolling shutter effects. Enhancing the \perframe module with advanced techniques could improve accuracy for these unaddressed distortions, though possibly at the expense of processing speed. Future work could explore trade-offs to enhance both accuracy and speed.

Regarding evaluation, we used the ScanNet dataset which primarily consists of diffuse objects, since our focus was on reconstructing opaque surfaces. For the future, extending the framework to handle transparent or reflective surfaces could be a promising research direction.

\section{Conclusion}

In this paper, we introduced \infusionsurf, a NeRF-style RGB-D 3D reconstruction framework that leverages \perframe and \tsdffusion to enhance reconstruction quality with minimal impact on optimization time.
The comprehensive comparative evaluations showed that our method is capable of accurately reconstructing a scene with high-frequency details.

\section{Acknowledgement}

This research was supported by the Korean Fund for Regenerative
Medicine (KFRM) grant funded by the Korea government (the
Ministry of Science and ICT, the Ministry of Health \& Welfare)
(23B0104L1) and by ROKIT Healthcare, Inc.


\title{Appendix}

\section{Appendix}

In this section, we present a series of experiments designed to demonstrate the effectiveness of our \perframe module. Additionally, we conduct a comparative study against other RGB-based 3D reconstruction methods to further demonstrate the strength of our method.

\subsection{Effectiveness of \perframe}
We performed a comprehensive study to evaluate the effectiveness of the \perframe using simulated errors in the intrinsic matrix on a synthetic dataset~\cite{NeuralRGBD}. To achieve this, we deliberately initialized the focal length to 570 instead of its ground truth value, which is 554.26. Additionally, we introduced random fluctuations following a normal distribution $\mathcal{N}(0,\,10^{2})$ in both the focal length and the principal points per frame. We then optimized our model, considering different scenarios by omitting either one or both of the \perframe and image plane deformation field (\cref{tab:ablation_quant}).
The quantitative result represented in \cref{tab:ablation_quant} implies that the image-plane deformation field and the \perframe impact complementarily on improving the quality of reconstructions.

\begin{table}[h!]
\centering
\begin{tabular}{cccccc}
\hline
\textbf{PFIR} & \textbf{IDPF} & \textbf{C-$\ell_1$} $\downarrow$         & \textbf{IoU} $\uparrow$            & \textbf{NC} $\uparrow$             & \textbf{F-Score} $\uparrow$        \\ \hline
              &               & 0.081          & 0.364          & 0.845          & 0.514          \\
              & \checkmark             & 0.067          & 0.486          & 0.861          & 0.645          \\
\checkmark             &               & 0.060          & 0.579          & 0.883          & 0.795          \\ \hline
\checkmark             & \checkmark            & \textbf{0.051} & \textbf{0.656} & \textbf{0.888} & \textbf{0.863} \\ \hline
\end{tabular}

\caption{Ablation study for the \perframe (PFIR) and image-plane deformation field (IPDF) on the quantitative dataset. Our PFIR and IPDF schemes demonstrate complementary impact on reconstruction quality.}
\label{tab:ablation_quant}

\end{table}

\subsection{Visualization of intrinsic-refined depth frame}
\begin{figure}[t!]
    \centering
    \rotatebox{90}{\makebox[0pt][c]{\small\hspace{5pt} Scene 2}}
    \begin{subfigure}[ht]{0.23\linewidth}
        \includegraphics[width=\linewidth]{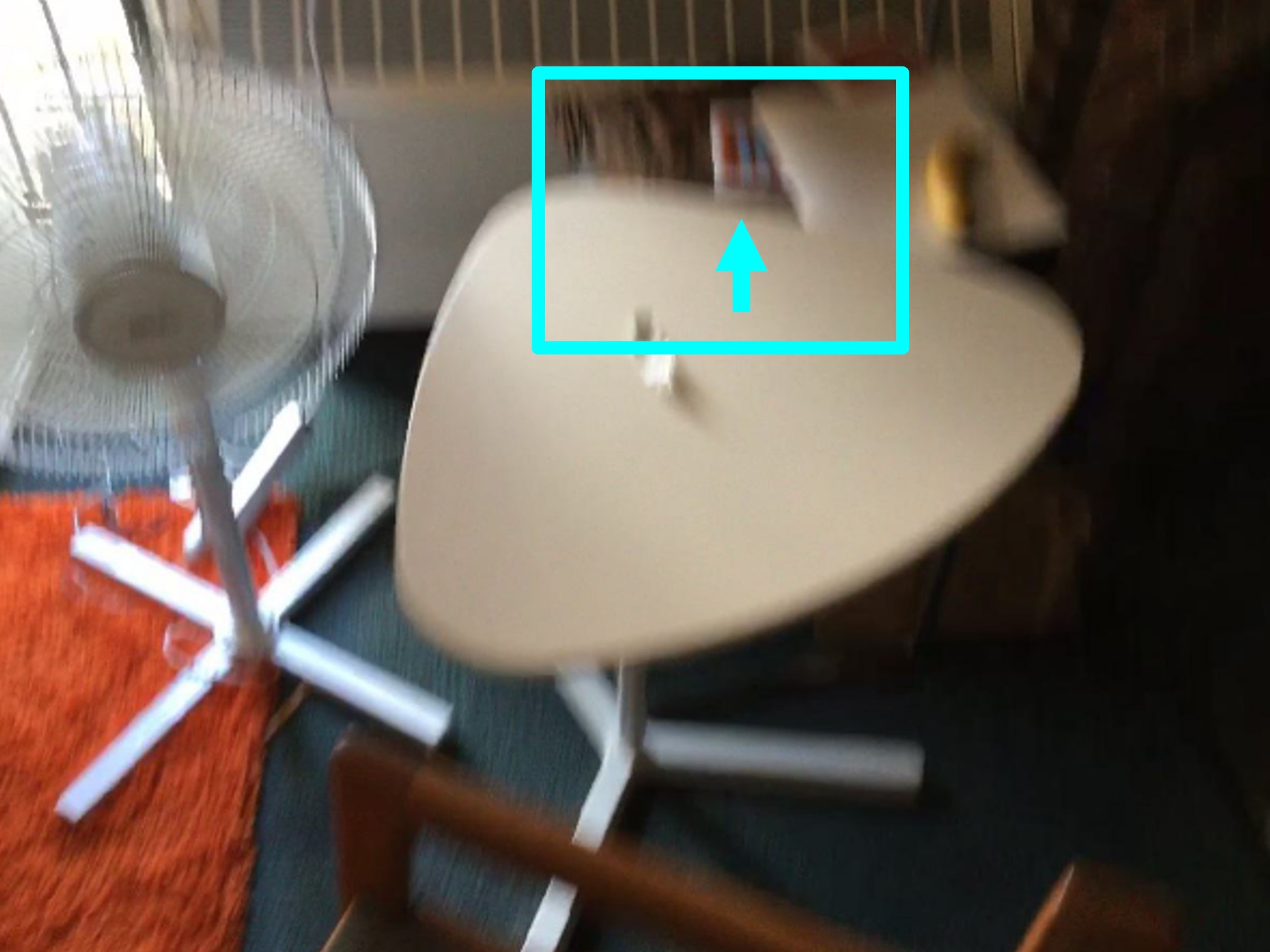}
        \centering
    \end{subfigure}
    \begin{subfigure}[ht]{0.23\linewidth}
        \includegraphics[width=\linewidth]{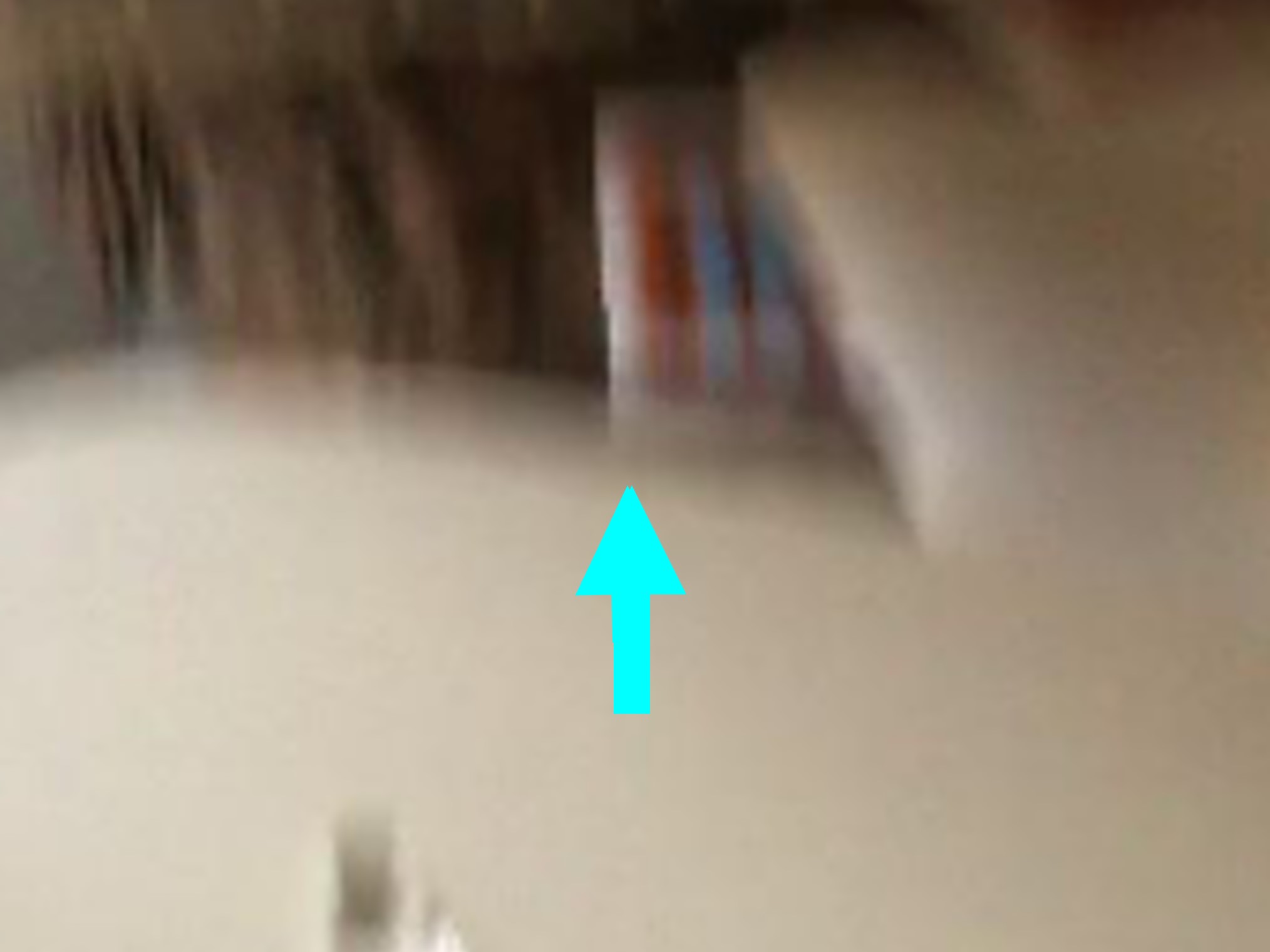}
        \centering
    \end{subfigure}
    \begin{subfigure}[ht]{0.23\linewidth}
        \includegraphics[width=\linewidth]{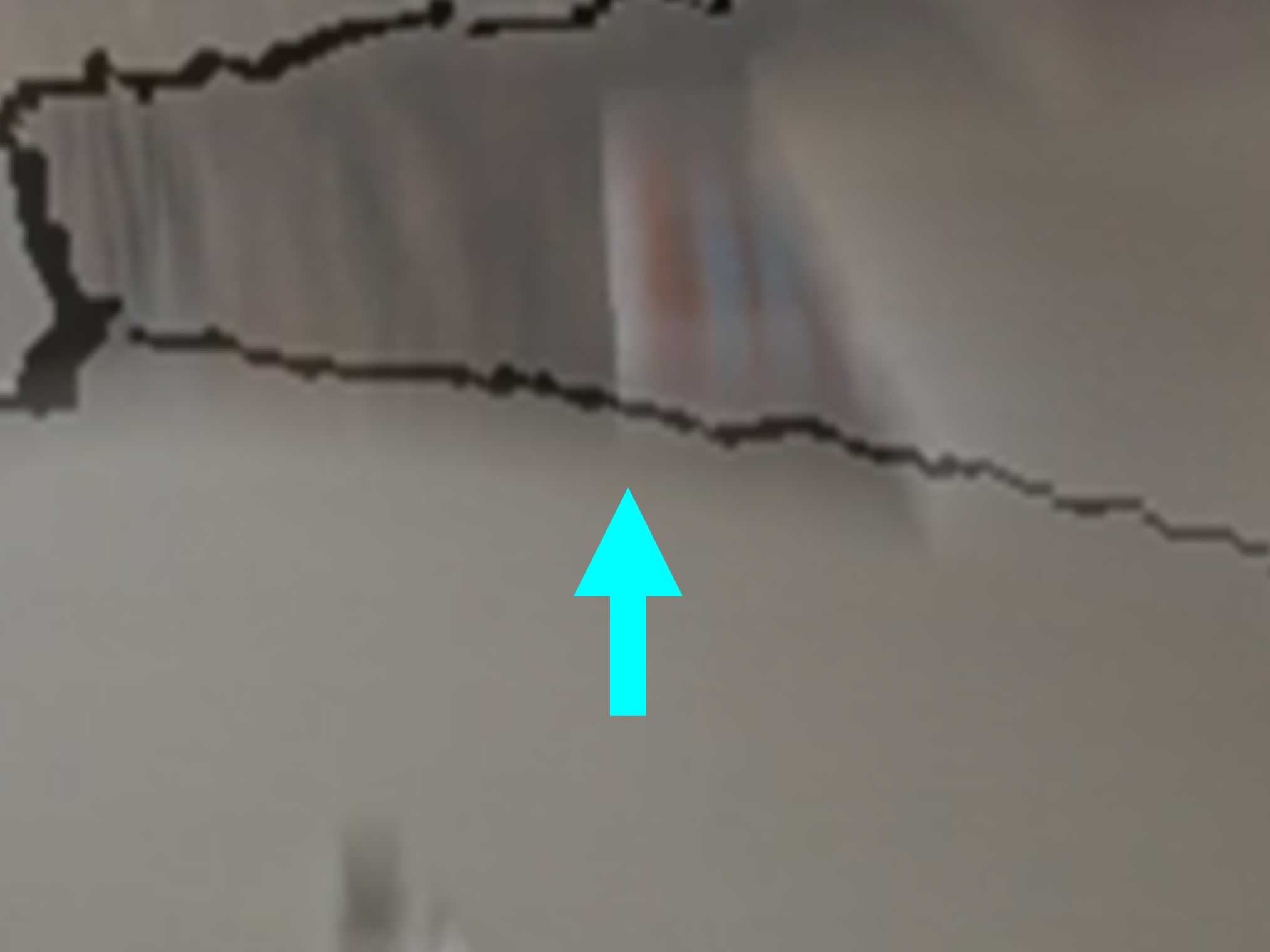}
        \centering
    \end{subfigure}
    \begin{subfigure}[ht]{0.23\linewidth}
        \includegraphics[width=\linewidth]{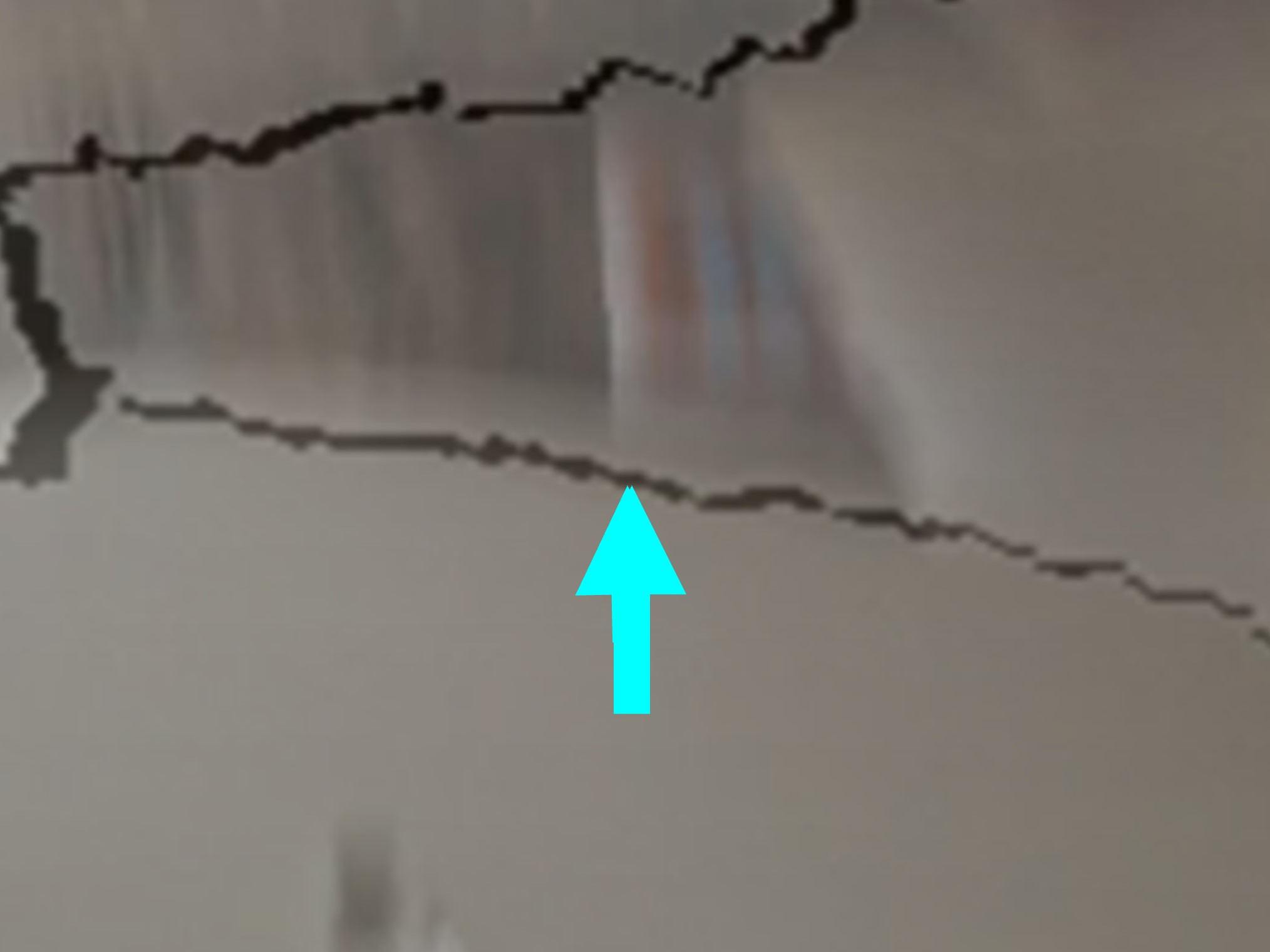}
        \centering
    \end{subfigure}
    \\
    \rotatebox{90}{\makebox[0pt][c]{\small\hspace{20pt} Scene 11}}
    \begin{subfigure}[ht]{0.23\linewidth}
        \includegraphics[width=\linewidth]{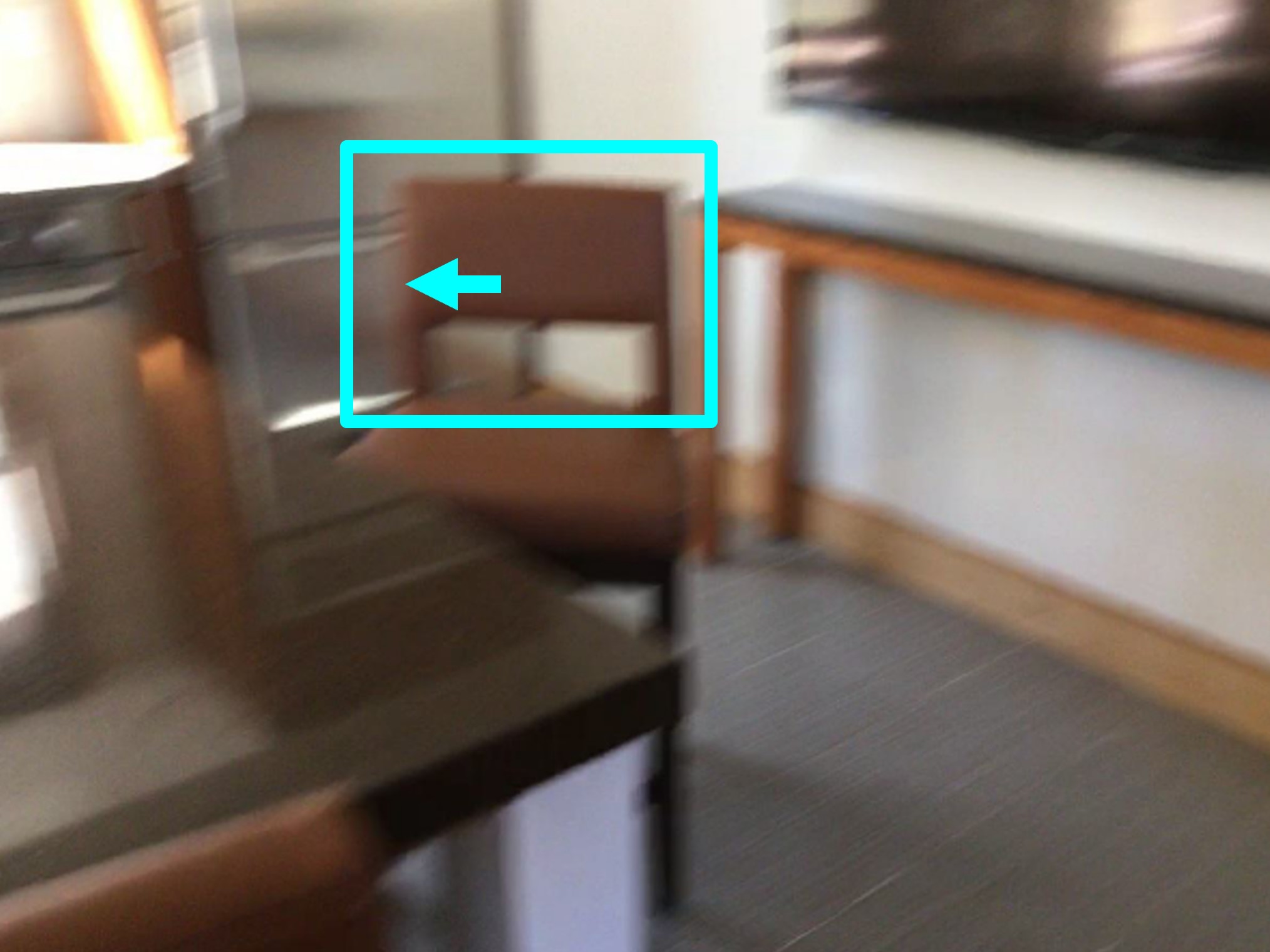}
        \centering
        \caption{}
        \label{subfig:refined_ori}
    \end{subfigure}
    \begin{subfigure}[ht]{0.23\linewidth}
        \includegraphics[width=\linewidth]{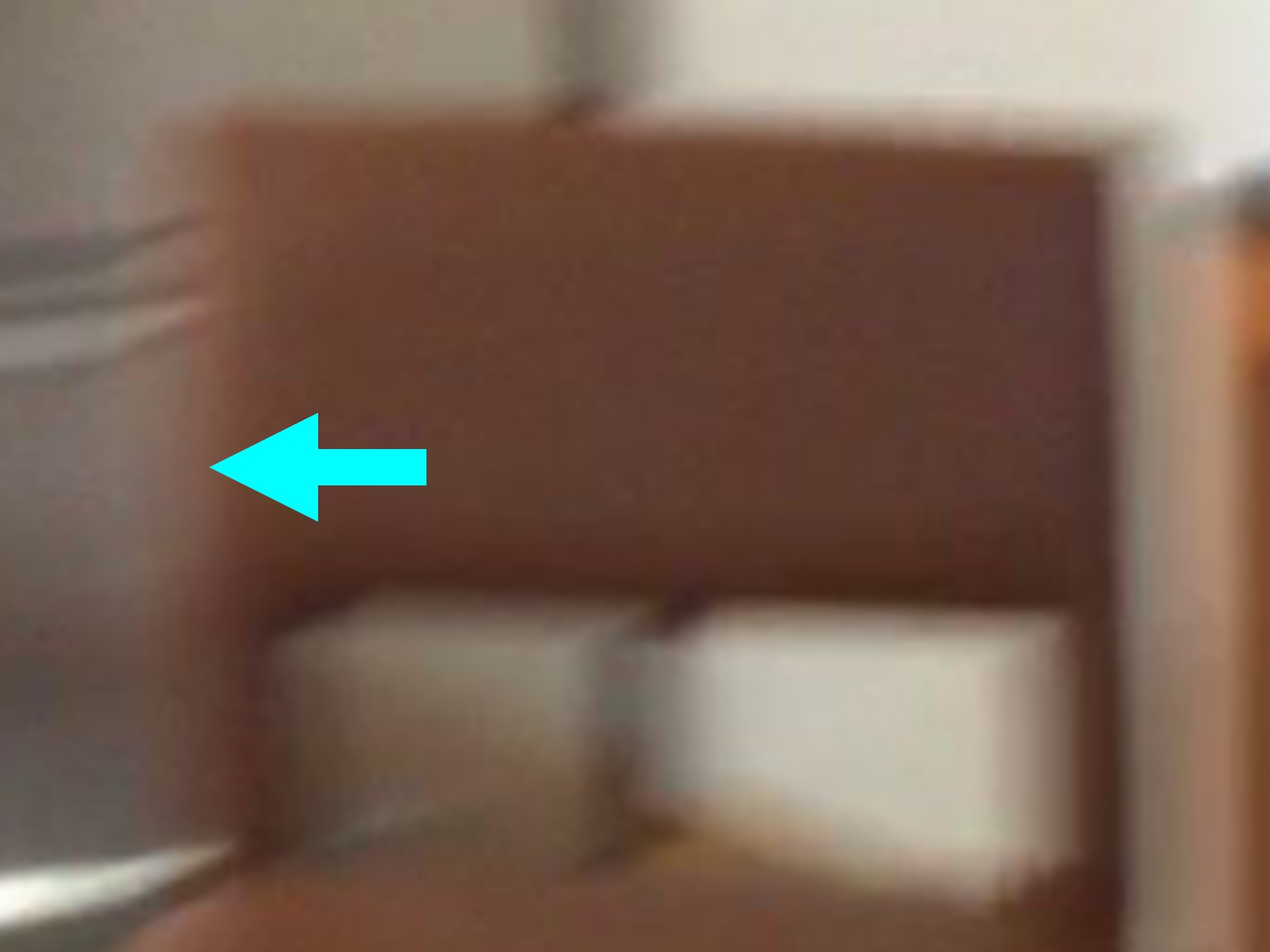}
        \centering
        \caption{}
        \label{subfig:refined_rgb}
    \end{subfigure}
    \begin{subfigure}[ht]{0.23\linewidth}
        \includegraphics[width=\linewidth]{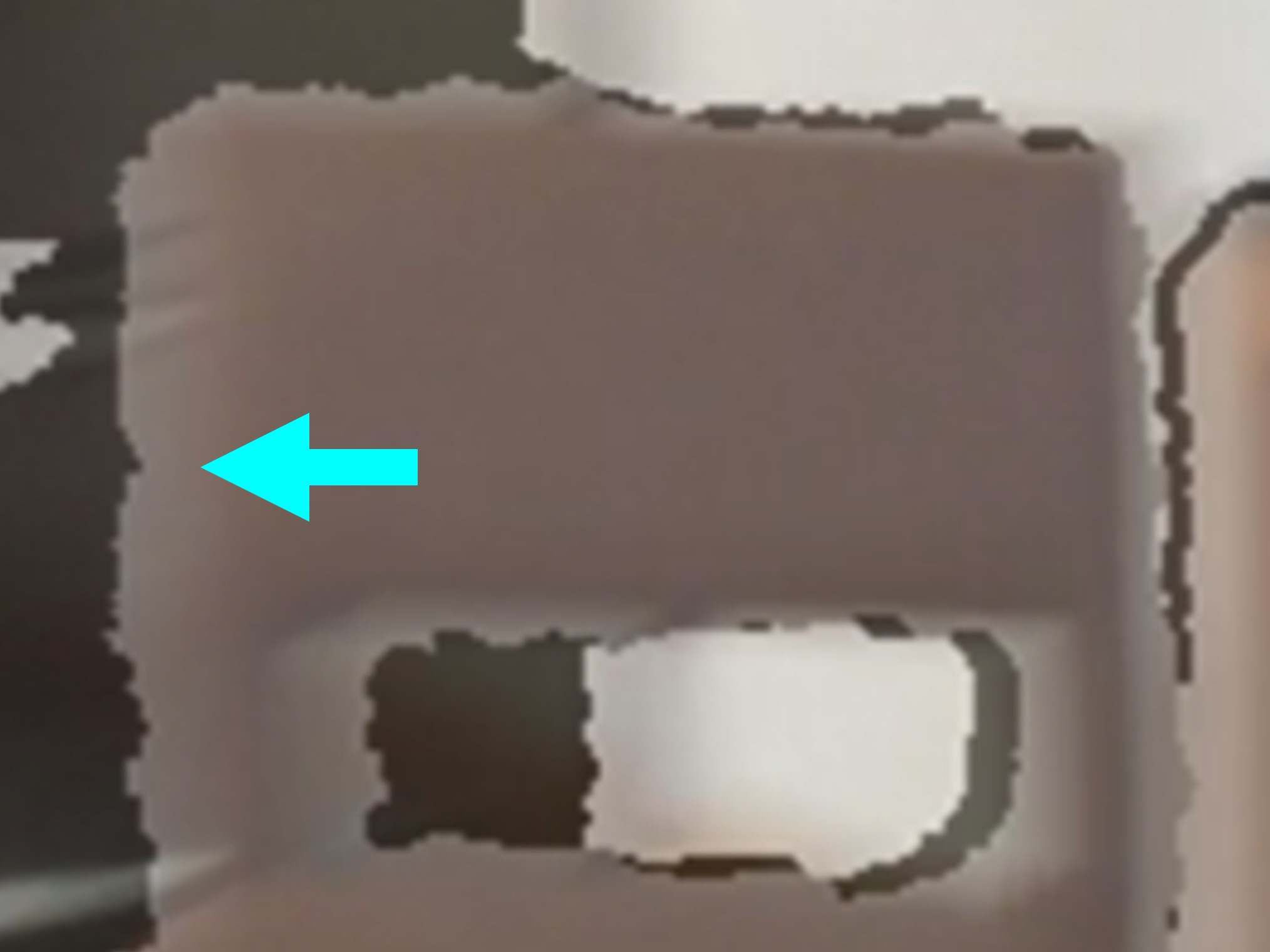}
        \centering
        \caption{}
        \label{subfig:refined_blur}
    \end{subfigure}
    \begin{subfigure}[ht]{0.23\linewidth}
        \includegraphics[width=\linewidth]{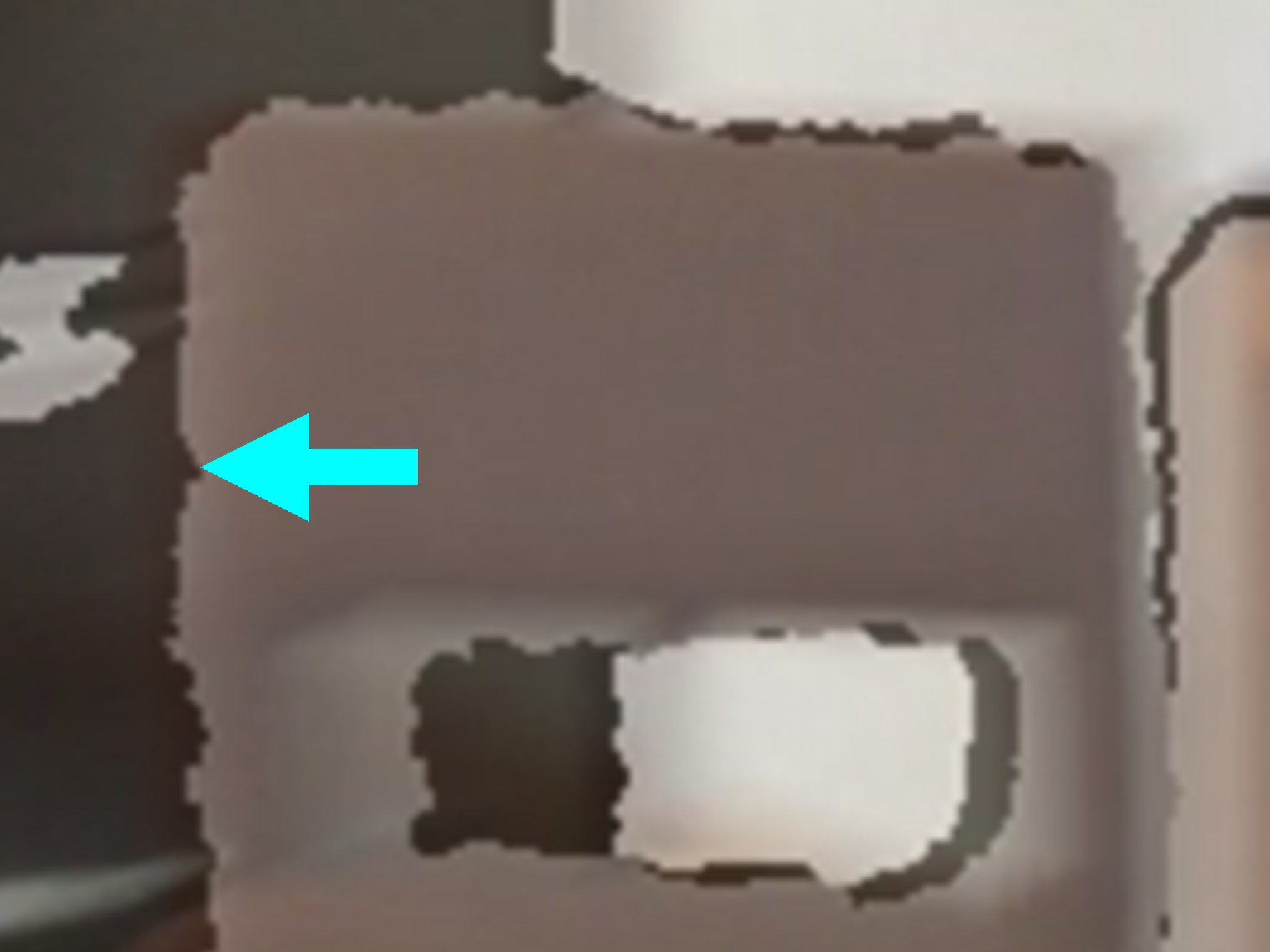}
        \centering
        \caption{}
        \label{subfig:refined_deblur}
    \end{subfigure}
    \\
    
    \caption{Visualization of our \perframe module. (a), (b) Color frames with camera motion blur. (c) Superimposed depth frames show incorrectly extended object boundaries. (d) Our \perframe module aligns the depth frames with the actual boundaries.}
    \label{fig:refined_images}
\end{figure}

To qualitatively evaluate the refinement results of our \perframe module, we performed experiments using depth frames from the ScanNet V2 dataset. For each frame, we transformed all pixels to image coordinates using the original intrinsic matrix, applied the scaling and translation and then re-projected them into pixels using the same intrinsic matrix.
As shown in \cref{fig:refined_images}, the results demonstrate the effectiveness of our \perframe module in correcting object boundaries within the depth frames. By compensating for errors caused by camera motion, our approach achieves more accurate reconstruction.

\subsection{Comparison against RGB-based approaches}

\begin{figure}[t]
    \centering
    \rotatebox{90}{\makebox[0pt][c]{\small\hspace{5pt} Scene 2}}
    \begin{subfigure}[ht]{0.3\linewidth}
        \includegraphics[width=\linewidth]{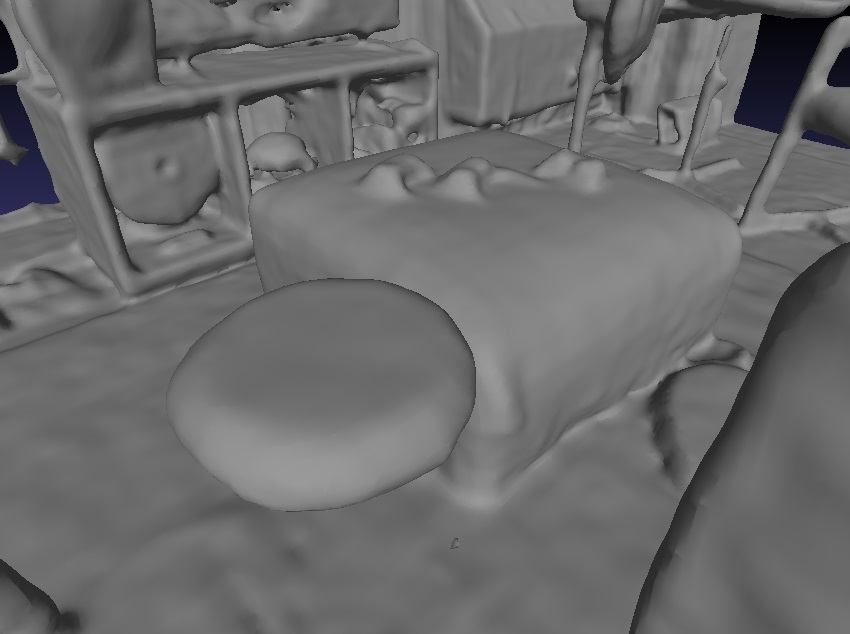}
        \centering
    \end{subfigure}
    \begin{subfigure}[ht]{0.3\linewidth}
        \includegraphics[width=\linewidth]{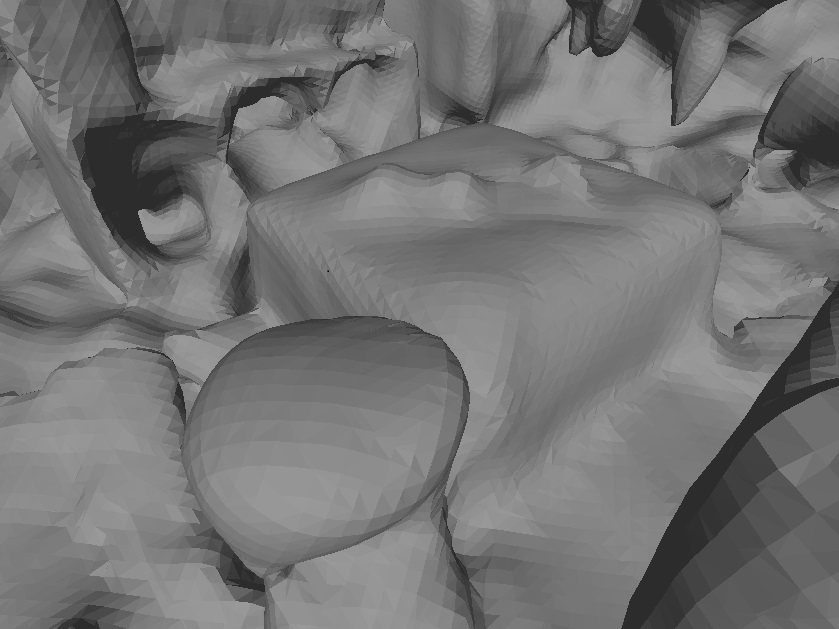}
        \centering
    \end{subfigure}
    \begin{subfigure}[ht]{0.3\linewidth}
        \includegraphics[width=\linewidth]{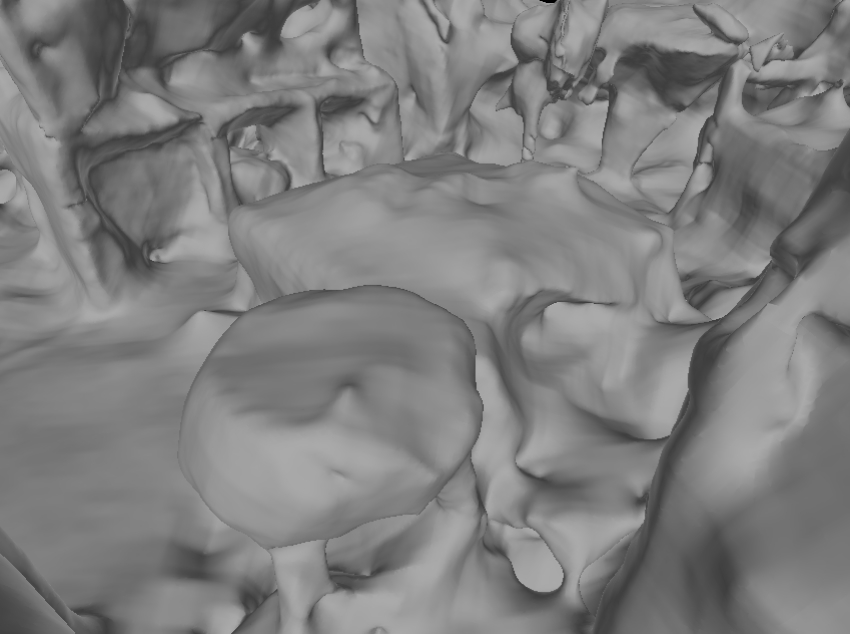}
        \centering
    \end{subfigure}
    \\
    \rotatebox{90}{\makebox[0pt][c]{\small\hspace{20pt} Scene 12}}
    \begin{subfigure}[ht]{0.3\linewidth}
        \includegraphics[width=\linewidth]{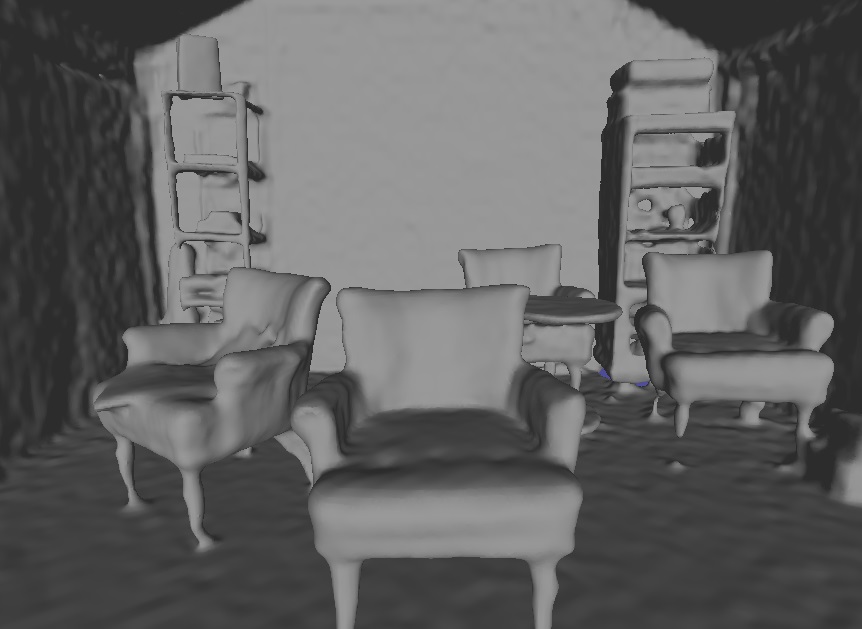}
        \caption{Ours}
        \centering
    \end{subfigure}
    \begin{subfigure}[ht]{0.3\linewidth}
        \includegraphics[width=\linewidth]{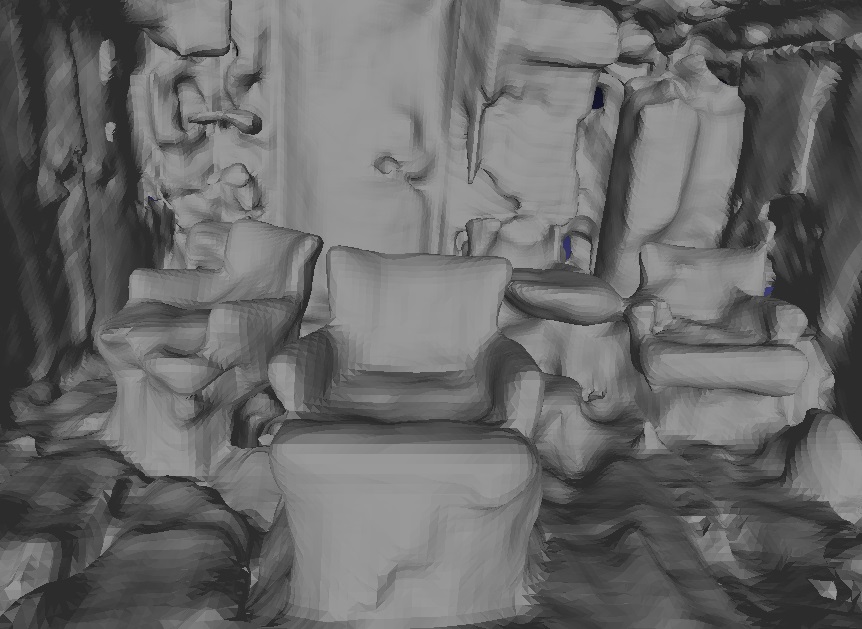}
        \caption{Voxurf~\cite{Voxurf}}
        \centering
    \end{subfigure}
    \begin{subfigure}[ht]{0.3\linewidth}
        \includegraphics[width=\linewidth]{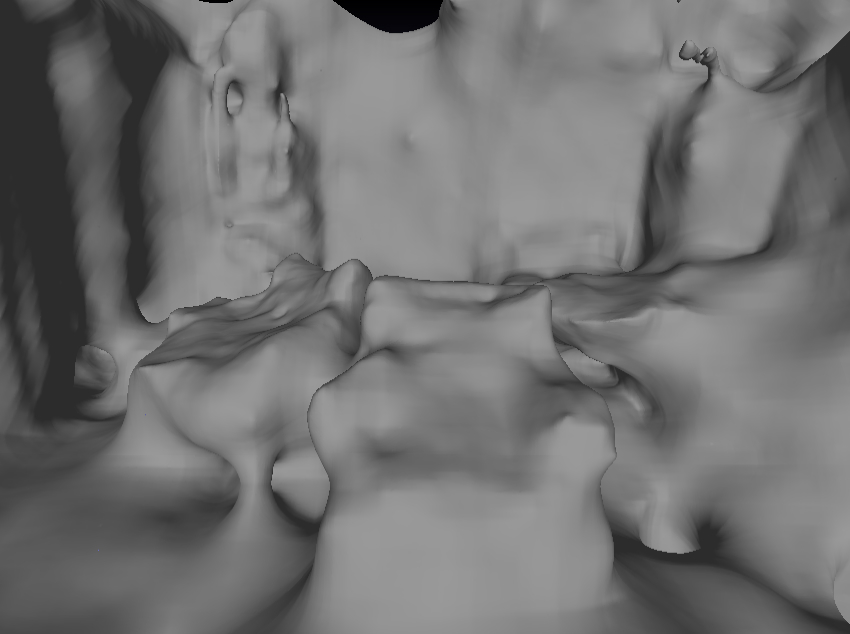}
        \caption{Neuralangelo~\cite{neuralangelo}}
        \centering
    \end{subfigure}
    \caption{Comparative result on indoor scenes from ScanNet V2~\cite{scannet}. This figure contrasts the effectiveness of our method (a) with that of RGB-based methods (b) and (c) in reconstructing the indoor scenes. The RGB-based methods appear to struggle with accurate reconstruction.}
    \label{fig:rgb_comp}
\end{figure}

Additionally, besides conducting comparative experiments with RGB-D based 3D reconstruction methods, we expanded our analysis to encompass two prominent RGB-based methods: Voxurf~\cite{Voxurf} and Neuralangelo~\cite{neuralangelo}. Employing COLMAP~\cite{colmap} with input RGB frames, we initially recovered camera parameters, then applied these methods to reconstruct indoor scenes from ScanNet V2~\cite{scannet}. As illustrated in Figure \ref{fig:rgb_comp}, RGB-based models struggled to accurately reconstruct the scenes.
We attribute this limitation to their inherent design, which relies on contextual information for multi-view constraints.
Particularly in sparse-view video capturing scenarios, relying solely on color input may not provide sufficient information for proper reconstruction.

\end{document}